\documentclass[11pt]{article}
\usepackage[utf8]{inputenc}
\usepackage[letterpaper,margin=1in]{geometry}

\usepackage{microtype}
\usepackage{graphicx}
\usepackage{subcaption}
\usepackage{booktabs} 

\usepackage{hyperref}

\usepackage{mathrsfs}
\usepackage{tabularx}
\usepackage{float}
\usepackage{caption}
\usepackage{amsmath,amssymb,amsfonts,bm}
\usepackage{color}
\usepackage{bbm}
\usepackage{comment}
\usepackage[textsize=tiny]{todonotes}
\usepackage{mathtools}
\usepackage{natbib}
\usepackage{empheq} 
\usepackage{commath}
\usepackage{appendix}
\usepackage{amsthm}
\usepackage{textcomp}
\usepackage[normalem]{ulem}
\usepackage{bm}
\usepackage{xcolor}
\usepackage{url}
\usepackage{authblk}
\usepackage[linesnumbered,ruled,vlined]{algorithm2e}

\usepackage{algorithmic}

\newtheorem{corollary}{Corollary}
\newtheorem{proposition}{Proposition}

\newcommand{\diag}{\mathrm{diag}}
\newcommand{\KL}{\mathrm{KL}}
\newcommand{\supp}{\mathrm{supp}}
\newcommand{\Ent}{\textsc{Ent}}
\DeclareMathOperator*{\argmax}{arg\,max}
\DeclareMathOperator*{\argmin}{arg\,min}
\DeclareMathOperator*{\tr}{tr}
\DeclareMathOperator{\Dkl}{D_{\KL}}

\title{Nonlinear Bayesian optimal experimental design using logarithmic Sobolev inequalities}

\author{Fengyi Li\thanks{fengyil@mit.edu}}
\author{Ayoub Belhadji\thanks{abelhadj@mit.edu}}
\author{Youssef Marzouk\thanks{ymarz@mit.edu}}

\affil{Massachusetts Institute of Technology}

\date{\today}

\begin{document}

\maketitle








\begin{abstract}
We study the problem of selecting $k$ experiments from a larger candidate pool, where the goal is to maximize mutual information (MI) between the selected subset and the underlying parameters. Finding the exact solution is to this combinatorial optimization problem is computationally costly, not only due to the complexity of the combinatorial search but also the difficulty of evaluating MI in nonlinear/non-Gaussian settings. We propose greedy approaches based on new computationally inexpensive lower bounds for MI, constructed via log-Sobolev inequalities. We demonstrate that our method outperforms random selection strategies, Gaussian approximations, and nested Monte Carlo (NMC) estimators of MI in various settings, including optimal design for nonlinear models with non-additive noise.

\end{abstract}



\section{Introduction}
The optimal experimental design (OED) problem arises in numerous settings, with applications ranging from combustion kinetics~\cite{huan_OEDnonlinear}, sensor placement for weather prediction~\cite{Krause08}, containment source identification~\cite{Attia17}, to pharmaceutical trials~\cite{djuris2024experimental}. A commonly addressed version of the OED problem centers on the fundamental question of selecting an optimal subset of $k$ observations from a total pool of $n$ possible candidates, with the goal of learning the parameters of a statistical model for the observations.

In the Bayesian framework, these parameters are endowed with a prior distribution to represent our state of knowledge before seeing the data. A posterior distribution on the parameters is obtained by conditioning on the observations. A commonly used experimental design criterion is then the mutual information (MI) between the parameters and the selected observations or, equivalently, the expected information gain from prior to posterior, which should be maximized. One na\"ive way to solve this problem is to enumerate all $\binom{n}{k}$ possible selections and choose the best $k$-subset. This problem is NP-hard~\cite{SFM_NPhard}; exhaustive enumeration is intractable even for moderate dimensions. Therefore, most efforts in the field are directed at finding approximate solutions. One widely used and straightforward approximation is to apply a greedy algorithm.

At each iteration, the standard greedy algorithm for cardinality-constrained maximization selects the observation that yields the largest improvement in the objective, conditioned on the observations that are already chosen, and appends this observation to the chosen set. However, this algorithm involves evaluating the incremental gain for each remaining observation at each iteration. This can be costly if evaluating the design objective is computationally expensive. For example, in our problem of interest, where MI is the objective, this process entails estimating MI $\mathcal O (nk)$ times. Estimating MI is understood to be challenging in the non-Gaussian setting~\cite{mcallester2020formal}.

One potential solution is to formulate a lower bound for the objective and to optimize over this bound rather than the original objective. \citet{iyer2012algorithms, narasimhan2012submodular} develop (sub)modular bounds and subsequently optimize over these bounds, for the purpose of minimizing the difference between two submodular functions. \citet{iyer2013fast} consider the problem of submodular function optimization by constructing bounds using discrete semidifferentials. \citet{jayanth} adopt a similar approach in the OED setting, and demonstrate performance comparable to the standard greedy approach. The bounds used in that work are specific to the linear-Gaussian problem, however, and not readily adapted to the  nonlinear/non-Gaussian setting. 

Separately, logarithmic Sobolev inequalities have been widely employed as a tool for linking the entropy of a function to another functional involving its gradient~\cite{log_sob_gross}. These inequalities have found application in dimension reduction~\cite{Zahm_CDR,Cui2021DatafreeLD}, the development of concentration inequalities~\cite{Boucheron2013ConcentrationI}, and analysis of the convergence of the Langevin samplers~\cite{chewi2021analysis, vempala2019rapid}. By selecting a particular form for the testing function and taking expectations, we can derive a tractable lower bound on the mutual information (equivalently, an upper bound on the difference between two conditional mutual informations~\cite{Zahm_CDR}). This approach enables optimization over the bound, rather than over the mutual information directly.

In this paper, we integrate the logarithm Sobolev inequality (LSI) into the standard greedy algorithm, presenting the ``logarithm Sobolev inequality greedy'' (LSIG) algorithm for nonlinear Bayesian OED. Our approach has several advantages over existing methods for design.
First, by optimizing the LSI bound, we circumvent the need to evaluate mutual information in the non-Gaussian setting, achieving significant computational savings. Directly estimating MI typically involves nested Monte Carlo (NMC) estimation, which converges more slowly than the standard Monte Carlo rate and requires large sample sizes~\cite{ryan2003estimating,huan_OEDnonlinear, rainforth2018nesting}; alternatively, evaluating a variational bound for mutual information involves training a neural network \citep{poole2019variational, foster_oed,belghazi2018mine,nguyen2010estimating,oord2018representation}, which again is far more expensive than our approach.
Additionally, the greedy iterations in LSIG are straightforward and cheap to implement. LSIG requires first calculating two matrices of interest, but then proceeds by selecting certain rows and columns from one, and taking Schur complements of blocks within the other---both of which are comparatively inexpensive operations. Choosing the design that maximizes the lower bound at each iteration is equivalent to selecting the largest element on the diagonal of the product of the two matrices. In contrast, the standard greedy approach necessitates iterating, at each stage, over all remaining designs.

\section{Problem setup and motivation}\label{sec:intro}
We consider the problem of selecting the $k$ most informative observations from $n>k$  candidates. Let $X\in \mathbb R^d$ be the parameter of interest and $Y \in \mathbb R^n$ be the vector of candidate observations. We use $A \subseteq \{1, \ldots, n\}$ to denote a set of indices and let $|A|$ denote its cardinality. Without loss of generality, we assume that the indices in $A$ are sorted in increasing order, and we use $A(i)$ to denote the $i$-th element in $A$. The complement of $A$, denoted by $\overline{A}$, is the set $\{1, \ldots, n\}\setminus  A$. We then use the notation $Y_A$ to represent the components of $Y$ whose indices are in $A$. For example, if $A = \{1,2,3\}$ and $n = 5$, then $Y_A = \{Y_1, Y_2, Y_3\}$ and $Y_{\overline{A}} = \{Y_4, Y_5\}$. To make analysis easier, we introduce the selection matrix $P_A\in \mathbb R^{n\times |A| }$ corresponding to $A$, with its $i$-th row being the $A(i)$-th row of the identity matrix of size $n\times n$. One can then write $Y_A = P_A^\top Y$. We hence seek the set $A$ such that the mutual information between $X$ and $Y_A$ is maximized, subject to the cardinality constraint $|A| = k$, i.e., 
\begin{align}
    A^* = \argmax_{A, |A|=k} I(X; Y_A). \label{Y_opt}
\end{align}
Here $I(\cdot, \cdot)$ denotes the mutual information between two random variables. Using the selection matrix notation, this is equivalent to maximizing $I(X; P_A^\top Y)$ over selection matrices $P_A$ of size $n \times k$ and obtaining the maximizer $P_{A^*}$.
%
For the rest of the paper, we use the notations $Y_A$ and $P_A^\top Y$ interchangeably depending on the context. 

Let $X$ and $Y$ have joint distribution $\pi_{X,Y}$ with marginals $\pi_X$ and $\pi_Y$. Using mutual information as a selection criterion lends itself to a natural Bayesian interpretation, wherein $\pi_X$ is regarded as the prior distribution. Moreover, we can write
\begin{align}
    I(X; Y_A) = \mathbb E_{Y_A}\left[\Dkl ( \pi_{X|Y_A}||\pi_{X} )\right].
\end{align}
That is, we are seeking a set of observations $Y_A$, or equivalent the set of indices $A$, such that the expected Kullback--Leibler (KL) divergence from the prior distribution of $X$ to the posterior distribution of $X$ given observations $Y_A$ is maximized. 
As noted in the introduction, one could solve this problem by brute-force enumeration, but this is intractable due to the number of possible subsets ${n\choose k}$. Even in the linear-Gaussian setting, where the MI objective has a closed-form expression, solving the problem by enumerating all feasible solutions is entirely impractical. We thus explore methods for obtaining approximate solutions.


\section{The standard greedy approach and challenges}
One natural approach of solving such problems is to consider  sequential selection. With a slight abuse of notation, let $A^k$ be any set of cardinality $k$. We then write 
\begin{align}
    I(X; Y_{A^k}) = \sum_{i = 0}^{k-1} I(X; Y_{A^{i+1}}) - I(X;Y_{A^{i}}).
\end{align}
To solve the original optimization problem, we can try to maximize each term inside the summation sign. Without introducing additional structure to the problem, this typically does not lead to the optimal solution; however, this notion guides us toward adopting a greedy approach, by solving each optimization problem inside the telescoping sum sequentially.
Suppose that we have already chosen $Y_{A^\ell}$ observations, with $\ell <k$, and we would like to find the next important observation. To be precise, at each iteration, we seek the index $i^*$ that satisfies the following:
\begin{align}\label{eq:greedy_gain}
    i^* = \argmax_{i\in\overline{A^\ell}} I(X; Y_{A^\ell \cup i}) - I(X;Y_{A^\ell}).
\end{align}
Therefore, during each stage, we are finding the observation that maximizes the incremental gain. Using the definition of conditional mutual information, this is exactly
\begin{align}
    i^* = \argmax_{i\in\overline{A^\ell}} I(X; Y_i|Y_{A^\ell}). \label{greedy}
\end{align}
Comparing \eqref{greedy} to \eqref{Y_opt}, we see that \eqref{greedy} is the conditioned version of \eqref{Y_opt}. In \eqref{Y_opt}, the set from which one can make selection is $\{1, \ldots, n\}$ and in \eqref{greedy}, the set becomes $\overline{A^\ell}$. Similar to the previous unconditioned setting, we can equivalently write \eqref{greedy} as a minimization problem,
\begin{align}\label{greedy_min}
     i^* = \argmin_{i\in\overline{A^\ell}} I(X; Y_{\overline{A^\ell}}|Y_{A^\ell}) - I(X; Y_i|Y_{A^\ell}),
\end{align}
by negating and adding a constant term. So far, we have obtained a formulation of the greedy approach. The critical question that now remains is how to efficiently solve \eqref{greedy_min}. The challenge here is two-fold. 
\begin{enumerate}
    \item One apparent challenge arises from the iterative and enumerative nature of the problem. We need to perform MI estimation for each $i\in\overline{A^\ell}$ at each iteration $\ell$. Suppose we would like to choose the best $k$ out of $n$ observations; then we need to perform mutual information estimation $\mathcal O (nk)$ times.

    
    \item Additionally, for each $i\in\overline{A^\ell}$, one needs to be able to estimate the mutual information between $X$ and $Y_A$. If $X$ and $Y$ are jointly normally distributed, then the mutual information admits a closed-form expression, which can be computed analytically. Outside of this setting, i.e., when the random variables are non-Gaussian, estimating the mutual information can be quite challenging, especially when the dimension of $Y_A$ is high. The standard approach to compute $I(X; Y_A)$ involves using NMC, which demands a substantial number of samples for accurate estimation. Another research avenue explores variational methods for approximating MI, creating and optimizing either upper or lower bounds (see Section~\ref{sec:related}). However, these methods necessitate training neural networks to obtain useful bounds, incurring significant computational costs. This becomes impractical when incorporated into each iteration of the standard greedy algorithm.

\end{enumerate} 

Given the aforementioned difficulties, we consider obtaining an approximate solution to \eqref{greedy_min}.
A frequently employed approach involves devising an upper bound for the minimization problem (or a lower bound for the maximization problem), followed by an attempt to optimize over this bound. The rationale behind this approach lies in the expectation that the optimization problem corresponding to constructed bound is easier to solve when compared to the original problem. We propose, in this paper, to use the LSI for constructing an upper bound to 
$I(X; Y_{\overline{A^\ell}}|Y_{A^\ell}) - I(X; Y_i|Y_{A^\ell})$. The constructed bound allows us to obtain an approximate solution to ~\eqref{greedy_min} by selecting the observation whose index corresponds to the largest diagonal entry of a particular diagnostic matrix.

\section{Constructing the bound}
In this section, we introduce an upper bound to $I(X; Y_{\overline{A^\ell}}|Y_{A^\ell}) - I(X; Y_i|Y_{A^\ell})$. As we have mentioned in the previous section, the optimization problem in each stage is a conditioned version of~\eqref{Y_opt}. Therefore, we first study the upper bound in the unconditioned setting and then extend it to the conditioned case. The upper bound is carefully constructed by exploiting the LSI.

\subsection{The log-Sobolev inequality}
\begin{proposition}[\citet{Ledoux_logsob,Guionnet2003,Zahm_CDR}]\label{prop:LSI}
    Let $\mu$ be a probability measure supported on a convex set with $\supp(\mu) \subseteq \mathbb R^p$. 
    We assume $\mu$ admits a Lebesgue density $\rho$ such that $\rho \propto \exp\left(-V- U \right)$, where $V\colon\supp(\mu) \rightarrow \mathbb R$ is a continuous twice differentiable function with $\nabla^2 V \succeq \Gamma^{-1}$ for some symmetric positive definite $\Gamma^{-1}$, and $U\colon\supp(\mu) \rightarrow \mathbb R$ is a bounded function such that $\exp(\sup U - \inf U) \leq \kappa$ for some $\kappa \geq 1$. Then $\mu$ satisfies 
    \begin{align}\label{eq:log_sob}
   \Ent_\mu\left[h^2\right]\leq 2\kappa \int \left\Vert \nabla h\right\Vert^2_{\Gamma} d\mu,
\end{align}
with the same log-Sobolev constant $\kappa$
for every continuously differentiable function $h$ defined on $\supp(\mu)$, where 
\begin{align*}
    \Ent_\mu\left[h^2\right] = \int h^2 \log \frac{h^2}{\int h^2 d\mu} d\mu
\end{align*}
is the entropy functional, and $\succeq$ is the Loewner order. Here, we call $\mu$ the reference measure. 
\end{proposition}
(Note that $\mu$ and the dimension $p$ are generic and may not necessarily be consistent with their usage elsewhere in this paper.) We would like to make several remarks here. First, for $\mu$ to satisfy a log-Sobolev inequality, it must be sub-Gaussian. Additionally, it is worth noting that $\kappa$ and $\Gamma$ are not unique. Specifically, if $\kappa$ and $\Gamma$ satisfy the LSI, then for any $\kappa' \geq \kappa$ and $\Gamma' \succeq \Gamma$, the pair $\kappa'$ and $\Gamma'$ also satisfy the LSI. This affords us some flexibility in selecting $\Gamma$ in the next part of this paper. Examples of measures $\mu$ that satisfy the LSI include Gaussians, Gaussian mixtures, and the uniform distribution on convex domains. More details can be found in~\citet{Zahm_CDR}.

Now, by choosing specific $\mu$ and $h$, we obtain the following corollary, whose proof is in Appendix~\ref{appendix:MI_LSI}
\begin{corollary}\label{cor:MI_LSI}
    Let $\mu = \pi_Y$, with $\supp(\mu)\subseteq \mathbb R^n$, satisfy the assumptions in Proposition ~\ref{prop:LSI}, and let the test function $h(y) = \sqrt{\frac{\pi_{X|Y}(x|y)}{\int \pi_{X|Y}(x|y) d\pi_Y(y)}}$ for a fixed $x$. Then we have 
    \begin{align}\label{eq:MI_bound}
    I(X;Y) \leq \frac{\kappa}{2} \int \left\Vert \nabla \log \pi_{X|Y}(x|y)\right\Vert^2_{\Gamma} d\pi_{X,Y}.
    \end{align}
    Furthermore, for a fixed $z$, let the reference measure $\mu_z = \pi_{Y|Z=z}$, with $\supp(\mu)\subseteq \mathbb R^n$, satisfy the assumptions in Proposition~\ref{prop:LSI}, and let the test function $h_z(y) = \sqrt{\frac{\pi_{X|Y, Z}(x|y, z)}{\int \pi_{X|Y, Z}(x|y, z) d\pi_{Y|Z}(y|z)}}$ for a fixed $x$ and $z$. Then 
    \begin{align}\label{eq:cond_MI_bound}
    I(X;Y|Z) \leq \frac{\kappa}{2} \int\left\Vert \nabla_y \log \pi_{X|Y,Z}(x|y,z) \right\Vert^2_{\Gamma_{Y|z}} d\pi_{X,Y,Z},
    \end{align}
    where $\kappa$, $\Gamma$ and $\Gamma_{Y|z}$ are both chosen to satisfy the assumptions in Proposition~\ref{prop:LSI} respectively. Note that $\kappa$ in~\eqref{eq:MI_bound} and~\eqref{eq:cond_MI_bound} are not necessarily the same. 
\end{corollary}

Recall that in the $\ell$-th iteration of the greedy scheme, we need to find the minimizer of the following expression, 
\begin{align}\label{MI_diff}
    I(X; Y_{\overline{A^\ell}}|Y_{A^\ell}) - I(X; Y_i|Y_{A^\ell}) = I(X; Y_{\overline{A^\ell\cup i}}|Y_{A^\ell\cup i})
\end{align}
over $i\in\overline{A^\ell}$.~\eqref{MI_diff} follows directly from the identity of conditioned mutual information. In the following analysis, we omit the dependence on $\ell$ for simplicity. 

In order to apply Corollary~\ref{cor:MI_LSI}, we assume that the reference measure $\pi_{Y_{\overline{A\cup i}}|Y_{A \cup i} = y_{A \cup i}}$ supported in $\mathbb R^{\left\vert\overline{A\cup i}\right\vert}$ satisfies the assumptions in Proposition~\ref{prop:LSI}. We can then apply~\eqref{eq:cond_MI_bound} to~\eqref{MI_diff} to obtain
\begin{align}\label{eq:conditioned_LSI}
    & I(X; Y_{\overline{A\cup i}}|Y_{A\cup i}) \nonumber\\
    \leq &\frac{\kappa}{2} \left(\int\left\Vert \nabla_{y_{\overline{A\cup i}}} \log \pi_{X|Y}(x|y_{\overline{A\cup i}}, y_{A\cup i})\right\Vert^2_{\Gamma_{Y_{\overline{A\cup i}}|Y_{A\cup i} =y_{A\cup i} }}  d\pi_{Y_{\overline{A\cup i}}|X, Y_{A \cup i}} d\pi_{X, Y_{A \cup i}} \right) \nonumber\\
    = & \frac{\kappa}{2} \left(\int\left\Vert \nabla_{y_{\overline{A\cup i}}} \log \pi_{X|Y}(x|y_{\overline{A\cup i}}, y_{A\cup i})\right\Vert^2_{\Gamma_{Y_{\overline{A\cup i}}|Y_{A\cup i} =y_{A\cup i} }}d\pi_{X,Y} \right),
\end{align}
with properly selected $\Gamma_{Y_{\overline{A\cup i}}|Y_{A\cup i} =y_{A\cup i} }$ (refer to Section~\ref{sec:greedy_bound} for further discussion).

\subsection{Bounding the greedy objective in each iteration}\label{sec:greedy_bound}

Note that~\eqref{eq:conditioned_LSI} is a conditioned version of~\eqref{eq:cond_MI_bound}, where everything is conditioned on the fact that $Y_{A\cup i} = y_{A\cup i}$ has already been chosen. $\kappa$ is the log-Sobolev constant such that~\eqref{eq:cond_MI_bound} holds, and the value of $\kappa$ may vary from one iteration to another. In a linear-Gaussian model, it is possible to compute both the log-Sobolev constant and $\Gamma_{Y_{\overline{A\cup i}}|Y_{A\cup i} =y_{A\cup i} }$. The resulting computations yield $\kappa = 1$ and $\Gamma_{Y_{\overline{A\cup i}}|Y_{A\cup i} =y_{A\cup i}} = \Sigma_{Y_{\overline{A\cup i}}|Y_{A\cup i}}$, where the latter represents the conditional covariance matrix. (We will generally use $\Sigma$, with subscripts as needed, to denote a covariance matrix.) 
Nevertheless, in  general settings, the challenges associated with implementing \eqref{eq:conditioned_LSI} become apparent. First of all, the reference measure $\pi_{Y_{\overline{A\cup i}}|Y_{A\cup i} =y_{A\cup i} }$ is typically not readily available, which poses a significant challenge when attempting to estimate the log-Sobolev constant and derive the analytical expression for $\Gamma_{Y_{\overline{A\cup i}}|Y_{A\cup i} =y_{A\cup i} }$. Furthermore, even when equipped with a straightforward expression for $\Gamma_{Y_{\overline{A\cup i}}|Y_{A\cup i} =y_{A\cup i} }$, there remains the necessity of computing an outer expectation with respect to $\pi_{X,Y_{A\cup i}}$. This entails the utilization of Monte Carlo evaluation, which results in an escalation of computational expenses.
To address these issues, we replace $\Gamma_{Y_{\overline{A\cup i}}|Y_{A\cup i} =y_{A\cup i} }$ with a matrix derived from a Gaussian approximation. 
To be more precise, by treating $Y$ as a Gaussian random variable, the conditional covariance $\Sigma_{Y_{\overline{A\cup i}}|Y_{A\cup i} =y_{A\cup i} } = \Sigma_{Y_{\overline{A\cup i}}|Y_{A\cup i}}$ is independent of the realization $y_{A\cup i}$, and can be obtained using the Schur complement given $\Sigma_Y$. Therefore, with potentially a different $\kappa$, we can write~\eqref{eq:conditioned_LSI} as
\begin{align}\label{eq:conditioned_LSI_G_cov}
    & I(X; Y_{\overline{A\cup i}}|Y_{A\cup i}) \nonumber\\
    \leq &
     \frac{\kappa}{2}\int\left\Vert \nabla_{y_{\overline{A\cup i}}} \log \pi_{X|Y}(x|y_{\overline{A\cup i}}, y_{A\cup i})\right\Vert^2_{\Sigma_{Y_{\overline{A\cup i}}|Y_{A\cup i}}}d\pi_{X,Y},
\end{align}
which is then upper bounded by
\begin{align}\label{eq:conditioned_LSI_G_cov_schur}
     \frac{\kappa}{2}\int\left\Vert \nabla_{y_{\overline{A\cup i}}} \log \pi_{X|Y}(x|y_{\overline{A\cup i}}, y_{A\cup i})\right\Vert^2_{\Sigma_{Y_{\overline{A\cup i}}|Y_{A}}}d\pi_{X,Y},
\end{align}
where we have used the fact that $\Sigma_{Y_{\overline{A\cup i}}|Y_{A\cup i}} \preceq \Sigma_{Y_{\overline{A\cup i}}|Y_{A}}$. This can be verified by observing that $\Sigma_{Y_{\overline{A\cup i}}|Y_{A\cup i}} = \Sigma_{Y_{\overline{A\cup i}}|Y_A} - \Sigma_{\left(Y_{\overline{A\cup i}}|Y_A\right)\left(Y_i|Y_A\right)}\Sigma_{Y_{i}|Y_A}^{-1} \Sigma_{\left(Y_i|Y_A\right)\left(Y_{\overline{A\cup i}}|Y_A\right)}$ with the Gaussian assumption, where we use the notation $\Sigma_{\left(Y_{\overline{A\cup i}}|Y_A\right)\left(Y_i|Y_A\right)}$ to denote the cross covariance between $Y_{\overline{A\cup i}}|Y_A$ and $Y_i|Y_A$. Expanding~\eqref{eq:conditioned_LSI_G_cov_schur} and using the cyclic property of the trace, we then obtain that 
\begin{align}\label{eq:conditioned_LSI_c}
   & \eqref{eq:conditioned_LSI_G_cov_schur} 
    = \frac{\kappa}{2} \tr \left(\mathbb E_{X,Y} \left[ \Sigma_{Y_{\overline{A\cup i}}|Y_{A}} \nabla_{y_{\overline{A\cup i}}} \log \pi_{X|Y}(x|y_{\overline{A\cup i}},y_{A\cup i}) \nabla_{y_{\overline{A\cup i}}} \log \pi_{X|Y}(x|y_{\overline{A\cup i}}, y_{A\cup i})^\top \right]\right) \nonumber\\
    &\leq \frac{\kappa}{2} \tr \left(\mathbb E_{X,Y} \left[  \Sigma_{Y_{\overline{A}}|Y_{A}} P_{\overline{A\cup i}|\overline{A}} P_{\overline{A\cup i}|\overline{A}}^\top  \nabla_{y_{\overline{A}}} \log \pi_{X|Y}(x|y)\nabla_{y_{\overline{A}}} \log \pi_{X|Y}(x|y)^\top \right]\right) + c,
\end{align}
where we use the notation $P_{\overline{A\cup i}|\overline{A}}$ to denote the selection matrix of choosing set $\overline{A\cup i}$ from $\overline{A}$, i.e., $P_{\overline{A\cup i}|\overline{A}}^\top \left(Y_{\overline{A}}|Y_{A} \right)= Y_{\overline{A\cup i}}|Y_{A}$, and hence the dimension of $P_{\overline{A\cup i}|\overline{A}}$ is $\left\vert\overline{A}\right\vert \times \left\vert\overline{A\cup i}\right\vert\ $. For the last inequality to hold, we add a constant $c$ to compensate for the fact that some diagonal elements of the matrix $\Sigma_{Y_{\overline{A}}|Y_{A}} P_{\overline{A\cup i}|\overline{A}} P_{\overline{A\cup i}|\overline{A}}^\top \nabla_{y_{\overline{A}}} \log \pi_{X|Y}(x|y) \nabla_{y_{\overline{A}}}\log \pi_{X|Y}(x|y)^\top$ might be negative. To be precise, let 
\begin{align*}
    \alpha = &\min \diag\left(\Sigma_{Y_{\overline{A}}|Y_{A}} \nabla_{y_{\overline{A}}} \log \pi_{X|Y}(x|y) \nabla_{y_{\overline{A}}} \log \pi_{X|Y}(x|y)^\top \right).
\end{align*}
We then set $c = -\alpha$, if $\alpha<0$ and $0$ otherwise. That is $c = \max \{-\alpha,0\}$. Now using the cyclic property of the trace one more time, we obtain that 
\begin{align*}
\eqref{eq:conditioned_LSI_c} 
    = & \tr \left(P_{\overline{A\cup i}|\overline{A}}^\top  \mathbb E_{X,Y} \left[   \nabla_{y_{\overline{A}}} \log \pi_{X|Y}(x|y) \nabla_{y_{\overline{A}}} \log \pi_{X|Y}(x|y)^\top  \right]\Sigma_{Y_{\overline{A}}|Y_{A}}P_{\overline{A\cup i}|\overline{A}}\right) + c. 
\end{align*}
Combing all the results, we obtain the log-Sobolev inequality for the conditional mutual information. That is, 
\begin{align}\label{eq:LSI_iter}
    &I(X; Y_{\overline{A\cup i}}|Y_{A\cup i}) \leq \frac{\kappa}{2}\tr \left(P_{\overline{A\cup i}|\overline{A}}^\top  \mathbb E_{X,Y} \left[   \nabla_{y_{\overline{A}}} \log \pi_{X|Y}(x|y) \nabla_{y_{\overline{A}}} \log \pi_{X|Y}(x|y)^\top  \right]\Sigma_{Y_{\overline{A}}|Y_{A}}P_{\overline{A\cup i}|\overline{A}}\right) + c.
\end{align}
It is crucial to emphasize that while the values of $\kappa$ and $c$ vary across iterations, they remain unrelated to the selection index $i$. Referring to ~\eqref{eq:LSI_iter}, in our pursuit to identify the minimizer of $I(X; Y_{\overline{A\cup i}}|Y_{A\cup i})$ with respect to the the index $i$, a practical approach would involve seeking an approximate solution by minimizing the right hand side. That is,
\begin{align}\label{eq:LSI_opt}
   i^* 
   & = \argmax_i \diag \left( \mathbb E_{X,Y} \left[   \nabla_{y_{\overline{A^\ell}}} \log \pi_{X|Y}(x|y) \nabla_{y_{\overline{A^\ell}}} \log \pi_{X|Y}(x|y)^\top  \right]\Sigma_{Y_{\overline{A^\ell}}|Y_{A^\ell}}, i\right),
\end{align}
where we use $\diag\left(A,i \right)$ to represent the $i$th entry on the diagonal of the matrix $A$. 
In each iteration, we require two specific matrices, $\mathbb E_{X,Y} \left[   \nabla_{y_{\overline{A^\ell}}} \log \pi_{X|Y}(x|y) \nabla_{y_{\overline{A^\ell}}} \log \pi_{X|Y}(x|y)^\top  \right]$ and $\Sigma_{Y_{\overline{A^\ell}}|Y_{A^\ell}}$. The former can obtained by calculating $\mathbb{E}_{X,Y} \left[ \nabla_y\log \pi_{X|Y}(x|y) \nabla_{y} \log \pi_{X|Y}(x|y)^\top \right]$ and selecting the rows and columns corresponding to $\overline{A^\ell}$, while the latter can be computed using Schur complement of the matrix block whose rows and columns are indexed by $A^\ell$. 
%

\subsection{Computing $\nabla_y \log \pi_{X|Y}(x|y)$}
One immediate question that arises is how should we compute $\nabla_y \log \pi_{X|Y}(x|y)$. We define its approximation as
\begin{align}\label{eq:grad_log_lik}
    \nabla_y \widehat{\log  \pi_{X|Y}}(x|y) = \nabla_y \log \pi_{Y|X}(y|x) - \nabla_y \widehat{\log \pi_Y}(y).
\end{align}
The first term on the right hand side is tractable if the likelihood function is differentiable with respect to $y$. For the second term, we approximate it using Monte Carlo. Observe that
\begin{align}\label{eq:MC_score}
    \nabla \widehat{\log \pi_Y}(y) \approx \nabla \log \pi_Y(y) = \frac{\nabla \pi_Y(y)}{\pi_Y(y)}\approx \frac{\nabla \widehat{\pi_Y}(y)}{\widehat{\pi_Y}(y)},
\end{align}
and we have
\begin{align}
    \widehat{\pi_Y}(y) &= \sum_{i = 1}^m \pi_{Y|X}(y|x^i) \approx \int \pi_{Y|X}(y|x) \pi_X(x) dx  = \pi_Y(y), \nonumber\\
      \nabla \widehat{\pi_Y}(y) &=\sum_{i = 1}^m \nabla_y \pi_{Y|X}(y|x^i)\approx \int \nabla_y \pi_{Y|X}(y|x) \pi_X(x) dx = \nabla \pi_Y(y) ,\label{eq:pi_y_MC}
\end{align}
where $\{x^i\}_{i=1}^m \sim \pi_X$. We summarize our proposed scheme in Algorithm \ref{alg:OED}. Another possible way of computing $\nabla \log \pi_Y(y)$ is to use score estimation~\cite{hyvarinen2005estimation,song2019generative,song2020sliced}. While these methods have demonstrated success in the realm of generative modeling, we do find that the relative magnitude of the elements of the matrix $\mathbb{E}_{X,Y} \left[ \nabla_y\log \pi_{X|Y}(x|y) \nabla_{y} \log \pi_{X|Y}(x|y)^\top \right]$ is quite sensitive when using neural networks to estimate the score function. Because we are concerned with the ranking of the diagonal elements of certain matrices, incorporating these aforementioned neural network-based methods into our framework appears challenging.

\begin{algorithm}[tb]
   \caption{LSIG for Bayesian OED}
   \label{alg:OED}
\begin{algorithmic}[1]
    \STATE {\bfseries Input}:  $A^0 = \O$,  samples from the joint distribution $\{x^i, y^i\}_{i=1}^M \sim \pi_{X,Y}$.
    \STATE {\bfseries Output:}  a set of indices $A^k$ of cardinality $k$
    \FOR {$i = 1$ to $M$}
    \STATE Compute $F^i = \nabla_y \widehat{\log  \pi_{X|Y}}(x^i|y^i)$ as in~\eqref{eq:grad_log_lik}.
    \ENDFOR
    \STATE Compute $F = \frac 1M\sum_{i = 1}^M F^i F^{{i}^\top} $
    \STATE Compute sample covariance matrix $S = \widehat\Sigma_Y$
    \FOR{$j = 1$ to $k$}
    \STATE $i^* = \argmax_i \diag \left( F S, i\right)$
    \STATE $A^j = A^{j-1} \bigcup i^*$.
    \STATE $F = F_{\overline{A^j}, \overline{A^j}}$ \hfill\COMMENT{$F_{\overline{A^j}, \overline{A^j}}$ refers to the submatrix with rows and columns indexed by the set $\overline{A^j}$}
    \STATE $S = S_{\overline{A^j}, \overline{A^j}} - S_{\overline{A^j}, A^j}S_{A^j, A^j}^{-1} S_{ A^j,\overline{A^j}}$ 
    \ENDFOR
 
\end{algorithmic}
\end{algorithm}


\section{Complexity}
In this section, we analyze the complexity of LSIG (Algorithm~\ref{alg:OED}), Gaussian approximation (see Appendix~\ref{appendix:GA}, Algorithm~\ref{alg:GA_standard_greedy}) and the NMC based greedy approach (NMC-greedy). For LSIG, suppose that we draw $m$ samples for estimating $\nabla \log \pi(y^i)$ for each $i = 1,\ldots, M$, and recycle these $m$ samples. For Gaussian approximation, we draw a total of $M$ samples for estimating the covariance matrices. For NMC-greedy, we solve~\eqref{eq:greedy_gain} in each iteration, with mutual information being computed using NMC with $M_{\mathrm{in}}$ inner loops and $ M_{\mathrm{out}}$ outer loops. The analysis is divided into two parts. We first analyze the number of sample draws and the number of model evaluations, where the latter is the number of times we evaluate $\pi_{Y|X}(y|x)$, $\nabla_y \log \pi_{Y|X}(y|x)$, or $\nabla_y \pi_{Y|X}(y|x)$. We then assess the complexity of linear algebra operations. The results are presented in Table~\ref{table:complexity}. We defer a detailed analysis to Appendix~\ref{appendix:complexity}. We observe that even when we set $M$, $m$, $M_{\mathrm{in}}$, and $M_{\mathrm{out}}$ to be of the same order of magnitude, NMC-greedy is far more expensive than LSIG. Moreover, practical use of NMC typically requires comparably much larger sample sizes, i.e., $M_{\mathrm{in}}M_{\mathrm{out}}$ much larger than $Mm$. 

\begin{table}

\caption{Complexity of LSIG, Gaussian approximation and NMC-greedy}
\begin{center}
\begin{tabular}{ cccc }
  \hline			
  &LSIG & \multicolumn{1}{p{1cm}}{\centering Gaussian \\ approximation} & NMC-greedy \\
  \hline
  Sample draws and model evaluations &$\mathcal O(Mm)$ & $\mathcal O(M)$ & $\mathcal O(nkM_{\mathrm{in}} M_{\mathrm{out}})$ \\
  \hline
  Linear algebra operations&$\mathcal O(nk^3)$ & $\mathcal O(nk^4)$ & $-$ \\
  \hline  
\end{tabular}
\label{table:complexity}

\end{center}
\end{table}

\section{Related work} \label{sec:related}
In the realm of combinatorial optimization under cardinality constraints, one research avenue focuses on studying the submodularity of the objective function.~\citet{das2011submodular, Iyer2013CurvatureAO, Bian17,liu2020submodular, jayanth} establish and study the concepts of curvature and submodularity ratio, with a focus on establishing theoretical guarantees for the performance of greedy algorithms. In contrast, our paper does not focus on optimization guarantees. Instead, we propose a \emph{new bound} that can be integrated into a greedy approach, with the aim of facilitating OED in nonlinear/non-Gaussian settings. 



A different line of research focuses on developing variational lower bounds for MI~\cite{poole2019variational, foster_oed,belghazi2018mine,nguyen2010estimating,oord2018representation}. Typically, however, the process of finding these bounds involves optimizing over the parameters of neural networks representing a so-called critic function, making these methods far more costly than our proposed scheme. For the purpose of OED, methods proposed by~\citet{foster2020unified,kleinegesse2021gradient,zhang2021scalable} simultaneously optimize over both the critic function and the design, enabling OED in a continuous domain. Although our method also optimizes over the bound rather than the original objective, there are fundamental differences. Firstly, the problem we are concerned with is discrete in nature---i.e., a combinatorial subset selection problem under cardinality constraints. This is distinct from selecting the best design on a continuous domain where \emph{gradient information} about the underlying design domain can be utilized. Additionally, the bound in our approach is obtained using a different technique that involves \textit{no} optimization or training.

\section{Numerical examples}
In this section, we present three numerical examples to illustrate our approach. The first one is a linear Gaussian problem, where the MI can be computed in a closed-form expression. The designs are selected using LSIG as well as the standard greedy. Note that in this case the standard greedy coincides with the Gaussian approximation. We then demonstrate our method on two nonlinear non-Gaussian examples, including an epidemic transmission model and a spatial Poisson model. Additional implementation details can be found in Appendix~\ref{appendix:numerical_examples}.

\subsection{The linear Gaussian problem}
In this section, we execute our proposed scheme alongside the standard greedy approach on a linear Gaussian problem. Since mutual information can be computed using a closed-form expression, it is convenient to compare the performance of the proposed method with that of the standard greedy. We consider the same model as in~\eqref{eq:linear_Gaussian}. Here $X \in \mathbb R^d, Y \in\mathbb R^{n}$. Consequently, $G \in \mathbb R^{d \times n}$ and $\epsilon \in \mathbb R^n$. Following Algorithm~\ref{alg:OED}, in each iteration, the problem we solve is 
\begin{align}\label{eq:LSIG_linear}
    i^* = \argmax_i \diag \left( \left(P^\top_{\overline{A}} \Sigma_\epsilon^{-1}P_{\overline{A}} \right) \Sigma_{Y_{\overline{A}}|Y_{A}}, i\right).
\end{align}
The detailed derivation of this optimization problem is presented in Appendix~\ref{appendix:linear_Gaussian}. For the standard greedy, on the other hand, we solve~\eqref{standard_greedy}, where we compute MI using the closed-form expression. Note that the Gaussian approximation (Algorithm~\ref{alg:GA_standard_greedy}) coincides with the standard greedy as this problem is linear and Gaussian. The forward model $G$, the covariance of the prior $\Sigma_X$ and the covariance of the noise $\Sigma_\epsilon$ are generated using exponential kernels (details can be read in Appendix~\ref{appendix:linear_Gaussian}), with spectrum shown in Figure~\ref{fig:LG_spectrum}. The dimension of $X$ and $Y$ are both set to be $50$. The MI is computed using the closed-form expression  $\frac 12 \log \det \frac{\Sigma_{Y_A}}{\Sigma_{\epsilon_A}}$ for a set $A$. The results are shown in Figure~\ref{fig:LG_performance}. As we can see, the results obtained using LSIG and the standard greedy (Gaussian approximation) are almost indistinguishable, whereas the results from $10$ random selection are noticeably worse.

It is worth noting that the selection criterion proposed in~\citet{jayanth} reads,
\begin{align}\label{eq:LSIG_linear_log}
    i^* = \argmax_i \diag \log\left( \left(P^\top_{\overline{A}} \Sigma_\epsilon^{-1} P_{\overline{A}}  \Sigma_{Y_{\overline{A}}|Y_{A}}\right), i\right),
\end{align}
which differs only by a matrix log factor. Although the optimal values of \eqref{eq:LSIG_linear} and~\eqref{eq:LSIG_linear_log} could be quite different, we observe numerically that they yield similar optimizers (note that here the matrix logarithm does not guarantee the same optimizer). In fact,~\eqref{eq:LSIG_linear_log} can be obtained using the \emph{dimensional} log-Sobolev inequality~\cite{dimensionalLSI}.

\subsection{Epidemic transmission model}\label{sec:epidemic}
\begin{figure}
\centering
\includegraphics[width=0.5\textwidth]{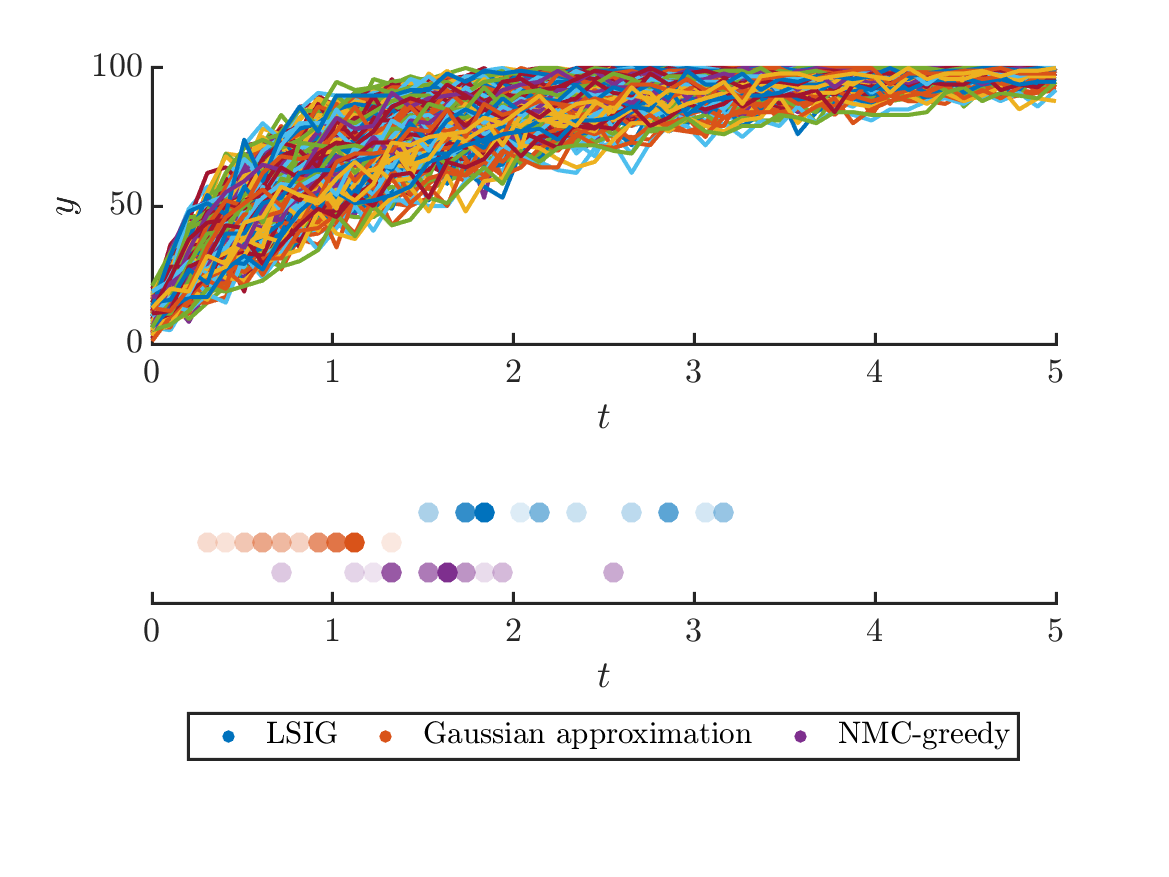}
\caption{Epidemic transmission model: (Top) The trajectory of the observations as a function of time. (Bottom) $10$ selected designs using different methods, with a darker color indicating the design being chosen in earlier stages. }
\label{fig:bd_realizations_designs}
\end{figure}

We study the model describing the transmission of epidemic~\cite{foster_oed,zhao2021inferencing}.  Given a population of size $N$, initially in a healthy state at time $t=0$. Individuals within this population undergo infection at a steady rate denoted by $x$ as time progresses. 
The prior is assumed to be log-normal, i.e., 
\begin{align*}
    \log X\sim \mathcal N(\mu_x, \sigma_x^2).
\end{align*}
We additionally discretize the time interval $[0, T_{\mathrm{end}}]$ by employing equispaced time points $t_1, \ldots, t_n$. The observation $y_i$  at time $t_i$ is modeled using negative binomial distribution:
\begin{align*}
    \pi_{Y_i|X}(y_i|x;t_i) = \frac{N!}{y_i!(N-y_i)!} \left(1-e^{-xt_i} \right)^{y_i} (e^{-xt_i})^{N-y_i}.
\end{align*}
We further assume that the observations $Y_i$ at each time are independent given $X$. Hence
\begin{align*}
     \pi_{Y|X}(y|x;t) =& \pi_{Y_1, \ldots, Y_n|X}(y_1, \ldots, y_n|X;t_1, \ldots, t_n) \\
     =& \prod_{i = 1}^n\pi_{Y_i|X}(y_i|x;t_i).
\end{align*}
For this problem, we set $N = 100$, $T_{\mathrm{end}}=5$, $n = 50$, $\mu_x = 0$ and $\sigma_x = 0.25$. The dimension of $X$ is $1$ in this case. The trajectory of the observations $y$ as a function of $t$ is shown in Figure~\ref{fig:bd_realizations_designs}. The objective is to identify the top 10 out of n time points in a way that maximizes the mutual information between the observations at those time points and the prior.

\begin{figure}
\centering
\includegraphics[width=0.4\textwidth]{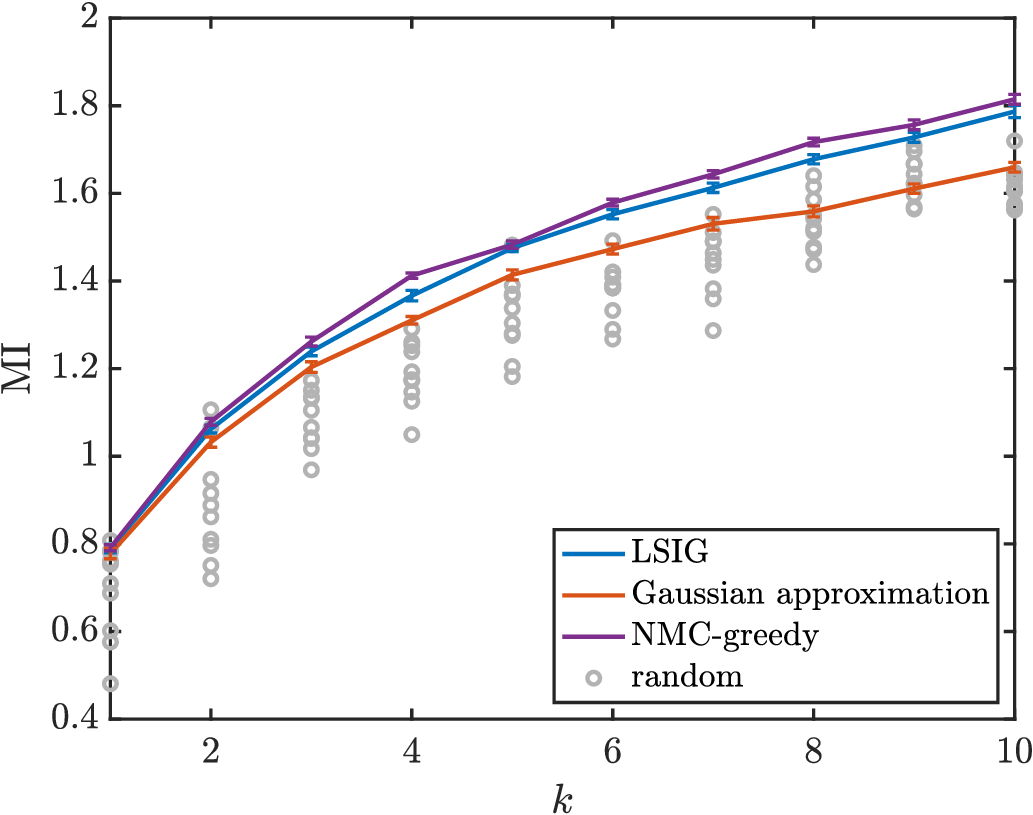}
\caption{Epidemic transmission model: MI for designs of increasing size, obtained using LSIG, NMC-greedy, and random selection. Error bars show one standard error, computed using $10$ trials. }
\label{fig:bd_MI}
\end{figure}

To implement Algorithm~\ref{alg:OED}, we need to compute $\nabla_y \log \pi_{Y|X}(y|x;t)$ and $\nabla_y \pi_{Y|X}(y|x;t)$ as~\eqref{eq:grad_log_lik} and~\eqref{eq:pi_y_MC}. The specific details for this computation are provided in the Appendix~\ref{appendix:epidemic}. As observed in Figure~\ref{fig:bd_MI}, the MI obtained through LSIG and the NMC-greedy approach exhibits similar performance, whereas the results computed using Gaussian approximation are noticeably inferior, being only comparable or slightly better than random selections. We also plot the selected designs in one instance in Figure~\ref{fig:bd_realizations_designs}. The designs chosen by LSIG and NMC-greedy differ fundamentally from those chosen using Gaussian approximation. Both LSIG and NMC-greedy select designs that lie in the middle, while the designs selected using Gaussian approximation appear to cluster in the region where the observations experience high variance.


\subsection{Spatial Poisson process}\label{sec:poisson}
We then study a two-dimensional non-homogeneous spatial Poisson model, which finds applications in wireless communication~\cite{keeler2014wireless} and particle systems~\cite{kostinski2000spatial,larsen2007spatial}. Now consider dividing a $[0,5]^2$ region into a grid of $25$ small cells with equal area. The number of events appearing in each cell follows a Poisson distribution, with intensity field being $b_iX_i$, where $b_i = l_i^{-1} D_i$, and $l_i$ denotes the distance from the center of the cell to the origin and $D_i$ is the area of the cell. We also have the prior 
\begin{align*}
    \log X\sim \mathcal N(0, \Sigma),\\
    \Sigma_{ij} = e^{-2\left\Vert c_i - c_j\right\Vert},
\end{align*}
where $c_i$ denotes the coordinate of the center of each cell in the grid. A realization of the intensity field can be seen in Figure~\ref{fig:poisson_intensity}. The likelihood model is defined as follows:
\begin{align*}
    \pi_{Y_i|X_i}(y_i|x_i; b_i) = \frac{(b_i x_i)^{y_i}}{y_i!}e^{-b_i x_i}.
\end{align*}
This expression represents the probability of observing $y_i$ events in the $i$-th cell. Assuming independence, we have 
\begin{align*}
    \pi_{Y|X}(y|x; b) = \prod_{i = 1}^n \pi_{Y_i|X_i}(y_i|x_i;b_i),
\end{align*}
and the goal is to select $10$ locations (designs), so that the mutual information between the observations in these locations and the prior $X$ is maximized. 

\begin{figure}
\centering
\includegraphics[width=0.4\textwidth]{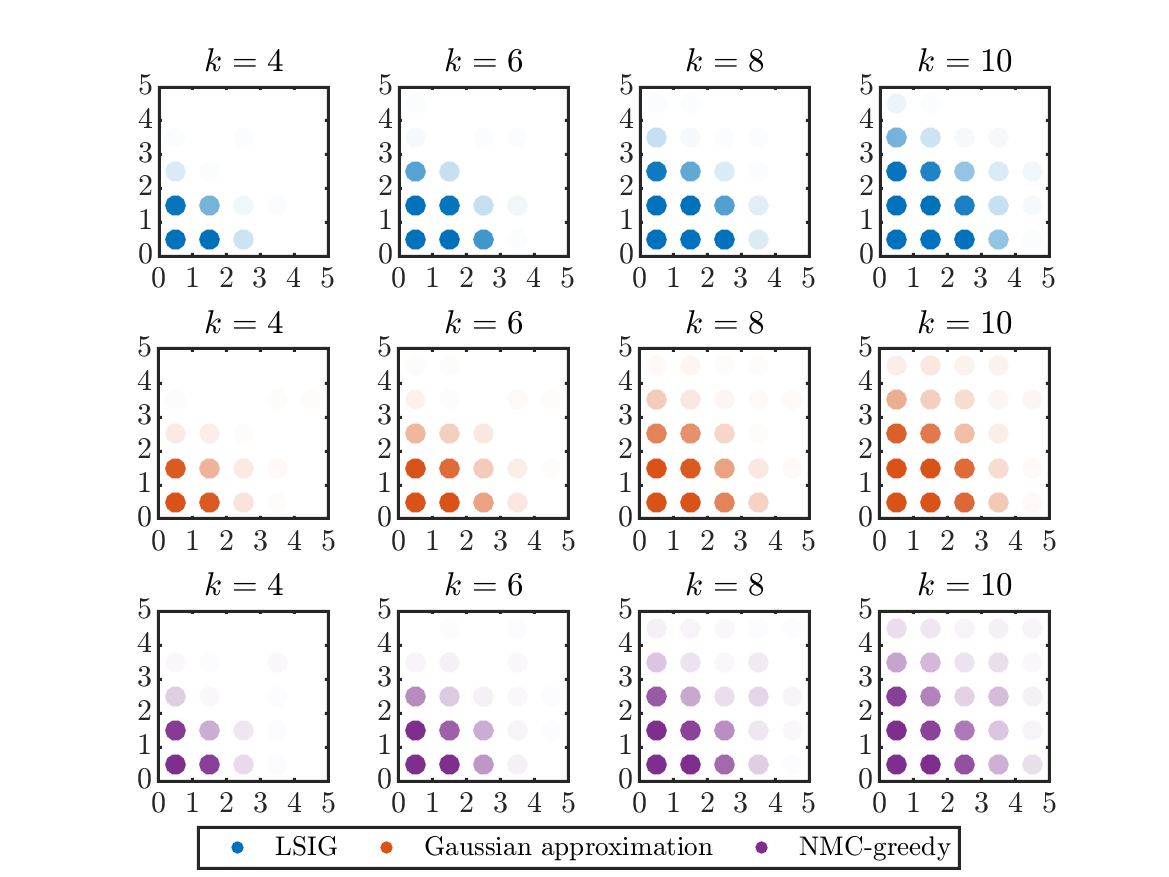}
\caption{Spatial Poisson process: The distribution of designs selected for $k = \{4,6,8,10\}$ using different methods, with darker colors indicating more frequent selections over $50$ repetitions.}
\label{fig:poisson_designs}
\end{figure}

\begin{figure}[h]
\centering
\includegraphics[width=0.4\textwidth]{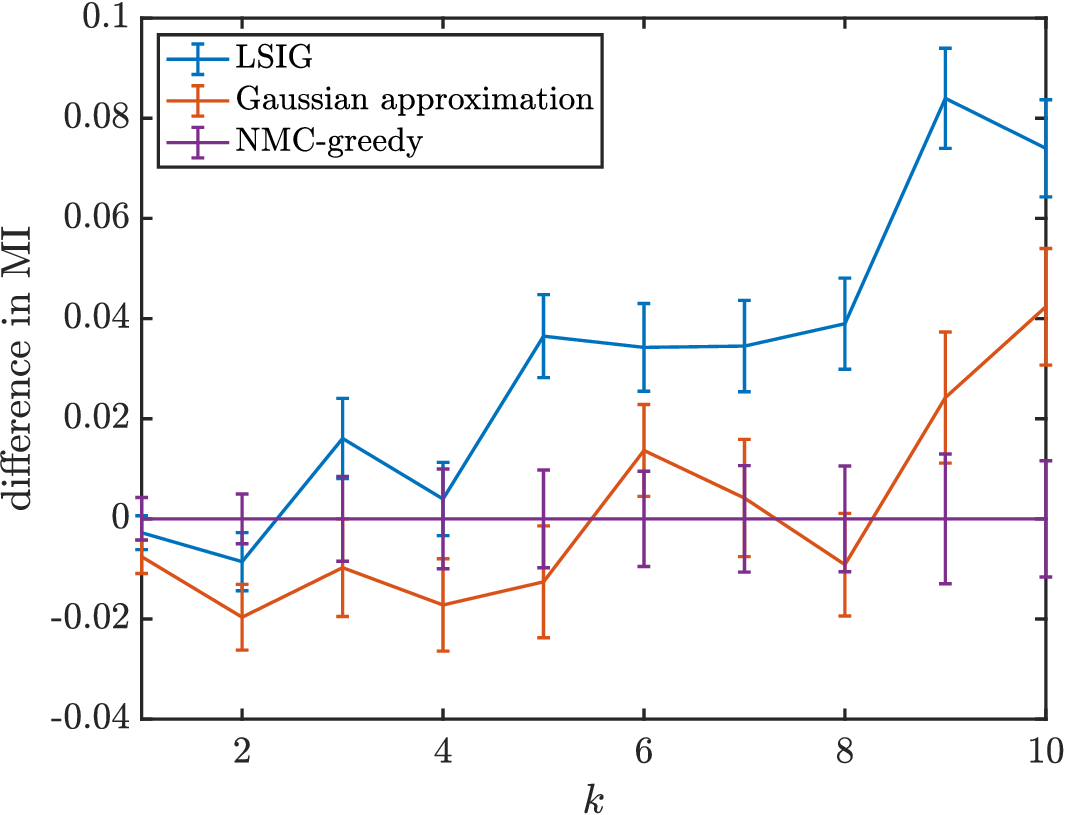}
\caption{Spatial Poisson process: differences in MI for designs obtained with \{LSIG, Gaussian approximation\} and with NMC-greedy (i.e., results obtained using NMC-greedy set the zero value at each $k$). MI values for the vertical axis are computed using NMC with $10000$ inner samples and $1000$ outer samples. Error bars show one standard error, computed over $50$ trials.}
\label{fig:poisson_MI_diff}
\end{figure}

 As in the previous example, we also need to compute $\nabla_y \log\pi_{Y|X}(y|x; b)$ and $\nabla_y \pi_{Y|X}(y|x;b)$. The detailed calculations are presented in Appendix~\ref{appendix:poisson}. For improved visualization, we standardize the results by subtracting the MI value obtained using NMC-greedy from that of LSIG and Gaussian approximation. (The plot with the performance of random selection is shown in Figure~\ref{fig:poisson_MI_diff_with_random}). As depicted in Figure~\ref{fig:poisson_MI_diff}, the MI obtained using LSIG outperforms that using the Gaussian approximation for all values of $k$. It also surpasses that obtained using NMC-greedy for all cases except $k = 1$ and $k = 2$. The error bars represent the standard error over $50$ trials. In Figure~\ref{fig:poisson_designs}, we show the distribution of designs for $k = \{4,6,8,10\}$ using different methods over $50$ trials, where darker colors indicate higher frequency of occurrence. We observe that the designs obtained using LSIG are more consistent and appeared to be less scattered compared to other methods. For example, for $k = \{4,6,8,10\}$, the number of cells being selected at least once over $50$ trials (i.e., the number of occupied cells) is $\{11,14,18,21\}$ using LSIG, while it is $\{14,18,22,23\}$ using Gaussian approximation, and $\{14,19,24,25\}$ using NMC-greedy, respectively.


\section{Conclusion}
We have introduced a new approach to solving the discrete nonlinear Bayesian OED problem. Specifically, we use log-Sobolev inequalities to develop new bounds for (conditional) MI that can be used within a greedy algorithm for combinatorial subset selection, eliminating the need to evaluate MI within each iteration. We demonstrate numerically that our method identifies designs of comparable quality to the standard greedy + MI estimation approach, but at much lower computational cost. 
In future work, we aim to extend our approach to the ``implicit model'' setting, where only samples are provided and the likelihood and its gradient are intractable, using tools for more consistent and stable score estimation.

\section{Acknowledgement}
The authors acknowledge support from the US Department of Energy, Office of Advanced Scientific Computing Research, award DE-SC0023188.

\bibliography{ref}

\begin{thebibliography}{40}
\providecommand{\natexlab}[1]{#1}
\providecommand{\url}[1]{\texttt{#1}}
\expandafter\ifx\csname urlstyle\endcsname\relax
  \providecommand{\doi}[1]{doi: #1}\else
  \providecommand{\doi}{doi: \begingroup \urlstyle{rm}\Url}\fi

\bibitem[Attia et~al.(2018)Attia, Alexanderian, and Saibaba]{Attia17}
Attia, A., Alexanderian, A., and Saibaba, A.~K.
\newblock Goal-oriented optimal design of experiments for large-scale bayesian linear inverse problems.
\newblock \emph{CoRR}, abs/1802.06517, 2018.
\newblock URL \url{http://arxiv.org/abs/1802.06517}.

\bibitem[Belghazi et~al.(2018)Belghazi, Baratin, Rajeswar, Ozair, Bengio, Courville, and Hjelm]{belghazi2018mine}
Belghazi, M.~I., Baratin, A., Rajeswar, S., Ozair, S., Bengio, Y., Courville, A., and Hjelm, R.~D.
\newblock Mine: mutual information neural estimation.
\newblock \emph{arXiv preprint arXiv:1801.04062}, 2018.

\bibitem[Bian et~al.(2017)Bian, Buhmann, Krause, and Tschiatschek]{Bian17}
Bian, A.~A., Buhmann, J.~M., Krause, A., and Tschiatschek, S.
\newblock Guarantees for greedy maximization of non-submodular functions with applications.
\newblock \emph{CoRR}, abs/1703.02100, 2017.
\newblock URL \url{http://arxiv.org/abs/1703.02100}.

\bibitem[Boucheron et~al.(2013)Boucheron, Lugosi, and Massart]{Boucheron2013ConcentrationI}
Boucheron, S., Lugosi, G., and Massart, P.
\newblock Concentration inequalities - a nonasymptotic theory of independence.
\newblock In \emph{Concentration Inequalities}, 2013.
\newblock URL \url{https://api.semanticscholar.org/CorpusID:124648595}.

\bibitem[Chewi et~al.(2021)Chewi, Erdogdu, Li, Shen, and Zhang]{chewi2021analysis}
Chewi, S., Erdogdu, M.~A., Li, M.~B., Shen, R., and Zhang, M.
\newblock Analysis of langevin monte carlo from poincar\'e to log-sobolev, 2021.

\bibitem[Cui \& Zahm(2021)Cui and Zahm]{Cui2021DatafreeLD}
Cui, T. and Zahm, O.
\newblock Data-free likelihood-informed dimension reduction of bayesian inverse problems.
\newblock \emph{Inverse Problems}, 37, 2021.
\newblock URL \url{https://api.semanticscholar.org/CorpusID:232068681}.

\bibitem[Cui et~al.(2024)Cui, Li, Li, Marzouk, and Zahm]{dimensionalLSI}
Cui, T., Li, F., Li, M., Marzouk, Y., and Zahm, O.
\newblock Tighter certified dimension reduction using dimensional logarithmic sobolev and poincare nequalities.
\newblock \emph{in preparation}, 2024.

\bibitem[Das \& Kempe(2011)Das and Kempe]{das2011submodular}
Das, A. and Kempe, D.
\newblock Submodular meets spectral: Greedy algorithms for subset selection, sparse approximation and dictionary selection.
\newblock \emph{arXiv preprint arXiv:1102.3975}, 2011.

\bibitem[Djuris et~al.(2024)Djuris, Vasiljevic, and Ibric]{djuris2024experimental}
Djuris, J., Vasiljevic, D., and Ibric, S.
\newblock Experimental design application and interpretation in pharmaceutical technology.
\newblock In \emph{Computer-Aided Applications in Pharmaceutical Technology}, pp.\  61--85. Elsevier, 2024.

\bibitem[Foster et~al.(2019)Foster, Jankowiak, Bingham, Horsfall, Teh, Rainforth, and Goodman]{foster_oed}
Foster, A., Jankowiak, M., Bingham, E., Horsfall, P., Teh, Y.~W., Rainforth, T., and Goodman, N.
\newblock \emph{Variational Bayesian Optimal Experimental Design}.
\newblock Curran Associates Inc., Red Hook, NY, USA, 2019.

\bibitem[Foster et~al.(2020)Foster, Jankowiak, O’Meara, Teh, and Rainforth]{foster2020unified}
Foster, A., Jankowiak, M., O’Meara, M., Teh, Y.~W., and Rainforth, T.
\newblock A unified stochastic gradient approach to designing bayesian-optimal experiments.
\newblock In \emph{International Conference on Artificial Intelligence and Statistics}, pp.\  2959--2969. PMLR, 2020.

\bibitem[Gross(1975)]{log_sob_gross}
Gross, L.
\newblock Logarithmic sobolev inequalities.
\newblock \emph{American Journal of Mathematics}, 97\penalty0 (4):\penalty0 1061--1083, 1975.
\newblock ISSN 00029327, 10806377.
\newblock URL \url{http://www.jstor.org/stable/2373688}.

\bibitem[Guionnet \& Zegarlinksi(2003)Guionnet and Zegarlinksi]{Guionnet2003}
Guionnet, A. and Zegarlinksi, B.
\newblock \emph{Lectures on Logarithmic Sobolev Inequalities}, pp.\  1--134.
\newblock Springer Berlin Heidelberg, Berlin, Heidelberg, 2003.
\newblock ISBN 978-3-540-36107-7.
\newblock \doi{10.1007/978-3-540-36107-7_1}.
\newblock URL \url{https://doi.org/10.1007/978-3-540-36107-7_1}.

\bibitem[Huan \& Marzouk(2013)Huan and Marzouk]{huan_OEDnonlinear}
Huan, X. and Marzouk, Y.~M.
\newblock Simulation-based optimal bayesian experimental design for nonlinear systems.
\newblock \emph{Journal of Computational Physics}, 232\penalty0 (1):\penalty0 288–317, 2013.
\newblock \doi{10.1016/j.jcp.2012.08.013}.

\bibitem[Hyv{\"a}rinen \& Dayan(2005)Hyv{\"a}rinen and Dayan]{hyvarinen2005estimation}
Hyv{\"a}rinen, A. and Dayan, P.
\newblock Estimation of non-normalized statistical models by score matching.
\newblock \emph{Journal of Machine Learning Research}, 6\penalty0 (4), 2005.

\bibitem[Iyer \& Bilmes(2012)Iyer and Bilmes]{iyer2012algorithms}
Iyer, R. and Bilmes, J.
\newblock Algorithms for approximate minimization of the difference between submodular functions, with applications.
\newblock \emph{arXiv preprint arXiv:1207.0560}, 2012.

\bibitem[Iyer et~al.(2013{\natexlab{a}})Iyer, Jegelka, and Bilmes]{iyer2013fast}
Iyer, R., Jegelka, S., and Bilmes, J.
\newblock Fast semidifferential-based submodular function optimization.
\newblock In \emph{International Conference on Machine Learning}, pp.\  855--863. PMLR, 2013{\natexlab{a}}.

\bibitem[Iyer et~al.(2013{\natexlab{b}})Iyer, Jegelka, and Bilmes]{Iyer2013CurvatureAO}
Iyer, R.~K., Jegelka, S., and Bilmes, J.~A.
\newblock Curvature and optimal algorithms for learning and minimizing submodular functions.
\newblock \emph{ArXiv}, abs/1311.2110, 2013{\natexlab{b}}.
\newblock URL \url{https://api.semanticscholar.org/CorpusID:8268885}.

\bibitem[Jagalur-Mohan \& Marzouk(2021)Jagalur-Mohan and Marzouk]{jayanth}
Jagalur-Mohan, J. and Marzouk, Y.
\newblock Batch greedy maximization of non-submodular functions: Guarantees and applications to experimental design.
\newblock \emph{J. Mach. Learn. Res.}, 22\penalty0 (1), jan 2021.
\newblock ISSN 1532-4435.

\bibitem[Keeler et~al.(2014)Keeler, Ross, and Xia]{keeler2014wireless}
Keeler, H.~P., Ross, N., and Xia, A.
\newblock When do wireless network signals appear poisson?, 2014.

\bibitem[Kleinegesse \& Gutmann(2021)Kleinegesse and Gutmann]{kleinegesse2021gradient}
Kleinegesse, S. and Gutmann, M.~U.
\newblock Gradient-based bayesian experimental design for implicit models using mutual information lower bounds.
\newblock \emph{arXiv preprint arXiv:2105.04379}, 2021.

\bibitem[Kostinski \& Jameson(2000)Kostinski and Jameson]{kostinski2000spatial}
Kostinski, A. and Jameson, A.
\newblock On the spatial distribution of cloud particles.
\newblock \emph{Journal of the atmospheric sciences}, 57\penalty0 (7):\penalty0 901--915, 2000.

\bibitem[Krause \& Golovin(2011)Krause and Golovin]{SFM_NPhard}
Krause, A. and Golovin, D.
\newblock Submodular function maximization.
\newblock \emph{Tractability}, 3:\penalty0 71--104, 01 2011.
\newblock \doi{10.1017/CBO9781139177801.004}.

\bibitem[Krause et~al.(2008)Krause, Singh, and Guestrin]{Krause08}
Krause, A., Singh, A., and Guestrin, C.
\newblock Near-optimal sensor placements in gaussian processes: Theory, efficient algorithms and empirical studies.
\newblock \emph{Journal of Machine Learning Research (JMLR)}, 9:\penalty0 235--284, February 2008.

\bibitem[Larsen(2007)]{larsen2007spatial}
Larsen, M.~L.
\newblock Spatial distributions of aerosol particles: Investigation of the poisson assumption.
\newblock \emph{Journal of aerosol science}, 38\penalty0 (8):\penalty0 807--822, 2007.

\bibitem[Ledoux(2000)]{Ledoux_logsob}
Ledoux, M.
\newblock Logarithmic sobolev inequalities for unbounded spin systems revisited.
\newblock 07 2000.
\newblock \doi{10.1007/978-3-540-44671-2_13}.

\bibitem[Liu et~al.(2020)Liu, Chong, Pezeshki, and Zhang]{liu2020submodular}
Liu, Y., Chong, E.~K., Pezeshki, A., and Zhang, Z.
\newblock Submodular optimization problems and greedy strategies: A survey.
\newblock \emph{Discrete Event Dynamic Systems}, 30:\penalty0 381--412, 2020.

\bibitem[McAllester \& Stratos(2020)McAllester and Stratos]{mcallester2020formal}
McAllester, D. and Stratos, K.
\newblock Formal limitations on the measurement of mutual information.
\newblock In \emph{International Conference on Artificial Intelligence and Statistics}, pp.\  875--884. PMLR, 2020.

\bibitem[Narasimhan \& Bilmes(2012)Narasimhan and Bilmes]{narasimhan2012submodular}
Narasimhan, M. and Bilmes, J.~A.
\newblock A submodular-supermodular procedure with applications to discriminative structure learning.
\newblock \emph{arXiv preprint arXiv:1207.1404}, 2012.

\bibitem[Nguyen et~al.(2010)Nguyen, Wainwright, and Jordan]{nguyen2010estimating}
Nguyen, X., Wainwright, M.~J., and Jordan, M.~I.
\newblock Estimating divergence functionals and the likelihood ratio by convex risk minimization.
\newblock \emph{IEEE Transactions on Information Theory}, 56\penalty0 (11):\penalty0 5847--5861, 2010.

\bibitem[Oord et~al.(2018)Oord, Li, and Vinyals]{oord2018representation}
Oord, A. v.~d., Li, Y., and Vinyals, O.
\newblock Representation learning with contrastive predictive coding.
\newblock \emph{arXiv preprint arXiv:1807.03748}, 2018.

\bibitem[Poole et~al.(2019)Poole, Ozair, Van Den~Oord, Alemi, and Tucker]{poole2019variational}
Poole, B., Ozair, S., Van Den~Oord, A., Alemi, A., and Tucker, G.
\newblock On variational bounds of mutual information.
\newblock In \emph{International Conference on Machine Learning}, pp.\  5171--5180. PMLR, 2019.

\bibitem[Rainforth et~al.(2018)Rainforth, Cornish, Yang, Warrington, and Wood]{rainforth2018nesting}
Rainforth, T., Cornish, R., Yang, H., Warrington, A., and Wood, F.
\newblock On nesting monte carlo estimators.
\newblock In \emph{International Conference on Machine Learning}, pp.\  4267--4276. PMLR, 2018.

\bibitem[Ryan(2003)]{ryan2003estimating}
Ryan, K.~J.
\newblock Estimating expected information gains for experimental designs with application to the random fatigue-limit model.
\newblock \emph{Journal of Computational and Graphical Statistics}, 12\penalty0 (3):\penalty0 585--603, 2003.

\bibitem[Song \& Ermon(2019)Song and Ermon]{song2019generative}
Song, Y. and Ermon, S.
\newblock Generative modeling by estimating gradients of the data distribution.
\newblock \emph{Advances in neural information processing systems}, 32, 2019.

\bibitem[Song et~al.(2020)Song, Garg, Shi, and Ermon]{song2020sliced}
Song, Y., Garg, S., Shi, J., and Ermon, S.
\newblock Sliced score matching: A scalable approach to density and score estimation.
\newblock In \emph{Uncertainty in Artificial Intelligence}, pp.\  574--584. PMLR, 2020.

\bibitem[Vempala \& Wibisono(2019)Vempala and Wibisono]{vempala2019rapid}
Vempala, S. and Wibisono, A.
\newblock Rapid convergence of the unadjusted langevin algorithm: Isoperimetry suffices.
\newblock \emph{Advances in neural information processing systems}, 32, 2019.

\bibitem[Zahm et~al.(2022)Zahm, Cui, Law, Spantini, and Marzouk]{Zahm_CDR}
Zahm, O., Cui, T., Law, K., Spantini, A., and Marzouk, Y.
\newblock Certified dimension reduction in nonlinear bayesian inverse problems.
\newblock \emph{Mathematics of Computation}, 91:\penalty0 1789--1835, 2022.
\newblock ISSN 0025-5718.
\newblock \doi{10.1090/MCOM/3737}.

\bibitem[Zhang et~al.(2021)Zhang, Bi, and Zhang]{zhang2021scalable}
Zhang, J., Bi, S., and Zhang, G.
\newblock A scalable gradient free method for bayesian experimental design with implicit models.
\newblock In \emph{International Conference on Artificial Intelligence and Statistics}, pp.\  3745--3753. PMLR, 2021.

\bibitem[Zhao et~al.(2021)Zhao, Shen, Musa, Guo, Ran, Peng, Zhao, Chong, He, and Wang]{zhao2021inferencing}
Zhao, S., Shen, M., Musa, S.~S., Guo, Z., Ran, J., Peng, Z., Zhao, Y., Chong, M.~K., He, D., and Wang, M.~H.
\newblock Inferencing superspreading potential using zero-truncated negative binomial model: exemplification with covid-19.
\newblock \emph{BMC Medical Research Methodology}, 21:\penalty0 1--8, 2021.

\end{thebibliography}
\bibliographystyle{icml2024}

\newpage
\appendix

\section{Proof of Corollary ~\ref{cor:MI_LSI}}\label{appendix:MI_LSI}
If we choose the reference measure $\mu = \pi_Y(y)$ and the test function $h(y) = \sqrt{\frac{\pi_{X|Y}(x|y)}{\int \pi_{X|Y}(x|y) d\pi_Y(y)}}$, then for a given $x$, we have 
\begin{align*}
    h^2(y) = \frac{\pi_{X|Y}(x|y)}{\pi_{X}(x)} = \frac{\pi_{Y|X}(y|x)}{\pi_{Y}(y)}.
\end{align*}
Then the left hand side of $\eqref{eq:log_sob}$ can be written as 
\begin{align*}
    \int \pi_{Y|X}(y|x) \log\frac{\pi_{Y|X}(y|x)}{\pi_{Y}(y)} dy = \Dkl(\pi_{Y|X}||\pi_{Y}),
\end{align*}
where we have used the property that $\int \frac{\pi_{Y|X}(y|x)}{\pi_{Y}(y)} d\pi_Y(y) = 1$. On the other hand, we have that 
\begin{align*}
    \nabla h(y) = \nabla_y\sqrt{\frac{\pi_{X|Y}(x|y)}{\pi_X(x)}} 
    = \frac {1}{2\sqrt{\pi_X(x)}} \frac{\nabla_y \pi_{X|Y}(x|y)}{\sqrt{\pi_{X|Y}(x|y)}},
\end{align*}
hence that
\begin{align*}
    \left\Vert \nabla h\right\Vert^2_{\Gamma} &= \frac 14 \left\Vert \nabla_y \log \pi_{X|Y}(x|y)\right\Vert^2_{\Gamma} \frac{\pi_{X|Y}(x|y)}{\pi_X(x)},
\end{align*}
and using the fact that $\frac{\pi_{X|Y}(x|y)}{\pi_X(x)} = \pi_{Y|X}(y|x)$, the right hand side of ~\eqref{eq:log_sob} becomes
\begin{align*}
    \frac{\kappa}{2} \int \left\Vert \nabla_y \log \pi_{X|Y}(x|y)\right\Vert^2_{\Gamma} \pi_{Y|X}(y|x) dy.
\end{align*}
Taking expectation with respect to $\pi_X$, we obtain
\begin{align*}
    \mathbb E_{\pi_X}\left[ \Dkl (\pi_{Y|X}||\pi_{Y}) \right] \leq \frac{\kappa}{2} \int \left\Vert \nabla \log \pi_{X|Y}(x|y)\right\Vert^2_{\Gamma} d\pi_{Y|X} d\pi_X.
\end{align*}
Note that $\mathbb E_{\pi_X}\left[ \Dkl (\pi_{Y|X}||\pi_{Y})  \right] = I(X;Y)$. Hence, 
\begin{align*}
    I(X;Y) \leq \frac{\kappa}{2} \int\left\Vert \nabla_y \log \pi_{X|Y}(x|y) \right\Vert^2_{\Gamma} d\pi_{X,Y}.
\end{align*}
We can further extend this to obtain an upper bound for conditional mutual information $I(X;Y | Z)$. Note that 
\begin{align*}
    I(X;Y|Z) = \mathbb E_{\pi_Z} \left[ \mathbb E_{\pi_{X|Z}}\left[ \Dkl (\pi_{Y|X,Z}||\pi_{Y|Z}) \right] \right] = \mathbb E_{\pi_{X,Z}} \left[ \Dkl (\pi_{Y|X,Z}||\pi_{Y|Z}) \right],
\end{align*}
and similarly, for a fixed $z$, we choose the reference measure $\mu_z = \pi_{Y|Z}(y|z)$ and test function $h_z(y) = \sqrt{\frac{\pi_{X|Y, Z}(x|y, z)}{\int \pi_{X|Y, Z}(x|y, z) d\pi_{Y|Z}(y|z)}}$. Hence $h_z^2(y) = \frac{\pi_{X|Y,Z}(x|y,z)}{\pi_{X|Z}(x|z)} = \frac{\pi_{Y|X,Z}(y|x,z)}{\pi_{Y|Z}(y|z)}$.
Then the left hand side of~\eqref{eq:log_sob} becomes
\begin{align*}
    \int \pi_{Y|X,Z}(y|x,z) \log\frac{\pi_{Y|X,Z}(y|x,z)}{\pi_{Y|Z}(y|z)} dy = \Dkl(\pi_{Y|X,Z}||\pi_{Y|Z}).
\end{align*}
Similarly, 
\begin{align*}
    \left\Vert \nabla h_z\right\Vert^2_{\Gamma_{Y|z}} &= \frac 14 \left\Vert \nabla_y \log \pi_{X|Y,Z}(x|y,z)\right\Vert^2_{\Gamma_{Y|z}} \frac{\pi_{X|Y,Z}(x|y,z)}{\pi_{X|Z}(x|z)},
\end{align*}
and right hand side of~\eqref{eq:log_sob} becomes

\begin{align*}
    \frac{\kappa}{2} \int \left\Vert \nabla_y \log \pi_{X|Y,Z}(x|y,z)\right\Vert^2_{\Gamma_{Y|z}} \pi_{Y|X,Z}(y|x,z) dy.
\end{align*}
Finally, taking expectation with respect to $\pi_{X,Z}$, we obtain
\begin{align*}
    I(X;Y|Z) \leq \frac{\kappa}{2} \int\left\Vert \nabla_y \log \pi_{X|Y,Z}(x|y,z) \right\Vert^2_{\Gamma_{Y|z}} d\pi_{X,Y,Z}.
\end{align*}

\section{The Gaussian approximation}\label{appendix:GA}

In this section, we consider simply, the Gaussian approximation to the problem. Given that we have already selected the set $A$, in the very next stage, the standard greedy method select the observation $i^*$ such that 
\begin{align}\label{standard_greedy}
    i^* = \argmax_{i\in \overline{A}} I(X;Y_{A\cup i}) - I(X;Y_{A}).
\end{align}
In the linear Gaussian case, mutual information can be computed in a closed form. Consider the following linear Gaussian problem,
\begin{align}\label{eq:linear_Gaussian}
    Y = GX + \epsilon,
\end{align}
where $X \sim \mathcal N(0, \Sigma_{X})$, $\epsilon \sim \mathcal N(0, \Sigma_\epsilon)$. In this case,~\eqref{standard_greedy} can be equivalently written as
\begin{align}\label{eq:GA}
    i^* = \argmax_{i\in \overline{A}} \log\det \Sigma_{Y_{A\cup i}} \Sigma_{\epsilon_{A\cup i}}^{-1},
\end{align}
given that the forward model is linear and the prior and the noise are both Gaussian.In the nonlinear setting, we can still apply this criterion by treating the random variables as if they were Gaussian and replacing the covariance matrix with its empirical counterpart. That is we consider solving the following problem in each iteration. 
\begin{align}
    i^* &= \argmax_{i\in \overline{A}} \log\det \left(P^\top_{A\cup i}\widehat\Sigma_{Y}P_{A\cup i} \right)\left(P^\top _{A\cup i}\widehat\Sigma_{Y|X}P _{A\cup i}\right)^{-1}\\
    &= \argmax_{i\in \overline{A}} \log\det \widehat\Sigma_{Y_{{A\cup i}}} \widehat\Sigma_{Y_{A\cup i}|X}^{-1},
\end{align}
where $\widehat\Sigma_{Y}$ can be estimated from samples and $\widehat\Sigma_{Y|X}$ can be obtained by first computing the joint covariance matrices $\widehat\Sigma_{X,Y}$, and then computing the Schur complement. We summarize the Gaussian approxiamtion in the algorithm \ref{alg:GA_standard_greedy}. 


\begin{algorithm}[tb]
   \caption{Gaussian approximation for Bayesian OED}
   \label{alg:GA_standard_greedy}
\begin{algorithmic}[1]
    \STATE {\bfseries Input}:  $A^0 = \O$,  samples from the joint distribution $\{x^i, y^i\}_{i=1}^M \sim \pi_{X,Y}$.
    \STATE {\bfseries Output:}  a set of indices $A^k$ of cardinality $k$
    \STATE Compute the sample covariance matrix $ \widehat\Sigma_Y$, and the conditional sample covariance matrix $\widehat\Sigma_{Y|X}$
    \FOR{$j = 1$ to $k$}
    \STATE $i^* = \argmax_{i\in \overline{A^{j-1}}}  \log\det \widehat\Sigma_{Y_{{A^{j-1}\cup i}}} \widehat\Sigma_{Y_{A^{j-1}\cup i}|X}^{-1}$.
    \STATE $A^j = A^{j-1} \bigcup i^*$.
    \ENDFOR
\end{algorithmic}
\end{algorithm}

\section{Analysis of complexity}\label{appendix:complexity}
We first analyze the number of samples draws and the number of times we call the likelihood function or the gradient of the likelihood function. For LSIG, suppose we use $m$ samples for estimating $\nabla \log \pi(y^i)$ for $i = 1,\ldots, M$, and recycle these $m$ samples. The total number of samples that we use is $m+M$. The number of function calls is calculated to be $M + 2Mm$. For Gaussian approximation, we need a total number of $M$ samples to construct the covariance matrices, and no function call is required. For NMC-greedy, implementing the outer greedy loop involves $\sum_{i=n-k+1}^n i = \frac{1}{2} k(2n-k+1)$ iterations. If we use $M_{\mathrm{in}}$ and $M_{\mathrm{out}}$ samples for the inner and outer loops for each MI computation using NMC respectively, then we require a total of $\frac{1}{2} k(2n-k+1) M_{\mathrm{in}} M_{\mathrm{out}}$ samples, as well as the same number of function calls. 

We then consider the computational complexity. Suppose we now have access to $\widehat\Sigma_Y$, $\widehat \Sigma_{Y|X}$ and an estimate of $\mathbb{E}_{X,Y} \left[ \nabla_y\log \pi_{X|Y}(x|y) \nabla_{y} \log \pi_{X|Y}(x|y)^\top \right]$. For LSIG, in the $i$-th iteration, computing the diagonal of the product of two matrices (the $9$th line in Algorithm~\ref{alg:OED}) has complexity of $\mathcal O\left((n-i+1)^2\right)$. Then, computing the Schur complement (the $12$th line in Algorithm~\ref{alg:OED}) involves matrix inversion, which has complexity of $\mathcal O((i-1)^3)$. Therefore, computing the Schur complement has complexity of $\mathcal O((n-i+1)(i-1)^2 + (i-1)^3)$. Summing it up from $i=1$ to $k$, the overall number of multiplications is dominated by $\mathcal O(nk^3)$. For the Gaussian approximation, in the $i$-th iteration, computing the determinant has a cost of $\mathcal O(i^3)$, and this has to be done for all the elements remains, resulting in a total cost of $\mathcal O(i^3 n)$. Summing from $i=1$ to $k$, the total complexity is dominated by $\mathcal O(nk^4)$.

\section{Numerical examples}\label{appendix:numerical_examples}
 In Section ~\ref{sec:epidemic} and~\ref{sec:poisson}, in order to compute $\nabla_y \widehat{\log\pi_{X|Y}}(x|y)$, we first generate $1000$ samples from the joint distribution, and for each $y^i$, where $i = 1,\ldots, 1000$, we calculate $\nabla \widehat{\pi_Y}(y^i)$ and $\widehat{\pi_Y}(y^i)$ using another set of $1000$ prior samples. To reduce computational costs, we recycle this batch of $1000$ prior samples and use them for all $i$. This results in a total number of $2000$ samples for implementing Algorithm~\ref{alg:OED}. To make fair comparison, we use a total number of $2000$ samples from the joint for implementing Algorithm~\ref{alg:GA_standard_greedy}. To obtain the results using NMC-greedy, we solve~\eqref{eq:greedy_gain} in each iteration, where the mutual information is computed using NMC, with inner and outer loops being $10000$ and $1000$ respectively. It it worth noting that if we set the total number of samples to be $2000$ for NMC-greedy, then, on average, the number of samples for computing MI is $5$ for~\ref{sec:epidemic} and $10$ for~\ref{sec:poisson}, rendering NMC infeasible. We also perform random selection, where in the $i$-th iteration, we randomly draw $i$ designs. After selecting the optimal designs using different methods, we then use nested Monte Carlo, still with inner and outer loops being $10000$ and $1000$, to compute the MI. This whole process is then repeated $10$ times for~\ref{sec:epidemic} and repeated $50$ times for~\ref{sec:poisson} in order to obtain reproducible results.  

\subsection{The linear Gaussian problem}\label{appendix:linear_Gaussian}
First note that the posterior $X|Y$ is also Gaussian, with mean $ \mu_{X|Y} = \Sigma_X G^\top \Sigma_Y^{-1} Y = \Sigma_{X|Y} G^\top \Sigma_{Y|X}^{-1} Y$, and covariance matrix $  \Sigma_{X|Y} = (\Sigma_X^{-1} + G^\top \Sigma_\epsilon^{-1} G)^{-1}$. 
From this we obtain that
\begin{align*}
    \nabla_y \log \pi_{X|Y = y}= \Sigma_{Y|X}^{-1} G (X -  \Sigma_{X|Y} G^\top \Sigma_{Y|X}^{-1} y).
\end{align*}
Then 
\begin{align*}
    \mathbb E_{X,Y} \left[   \nabla_{y} \log \pi_{X|Y}(x|y) \nabla_{y} \log \pi_{X|Y}(x|y)^\top  \right] = \Sigma_{Y|X}^{-1} G \Sigma_{X|Y} G^\top  \Sigma_{Y|X}^{-1} = \Sigma_\epsilon^{-1} - \Sigma_Y^{-1}.
\end{align*}
For the LSIG, at each iteration, the objective function is 
\begin{align*}
    P^\top_{\overline{A}} \left( \Sigma_\epsilon^{-1} - \Sigma_Y^{-1}\right)P_{\overline{A}} \Sigma_{Y_{\overline{A}}|Y_{A}}=
    \left(P^\top_{\overline{A}} \Sigma_\epsilon^{-1}P_{\overline{A}} \right) \Sigma_{Y_{\overline{A}}|Y_{A}}- I
\end{align*}
where we use the identity that 
\begin{align*}
    P^\top_{\overline{A}} \Sigma_Y^{-1}P_{\overline{A}}  = \Sigma_{Y_{\overline{A}}|Y_{A}}^{-1}.
\end{align*}
Therefore, in each iteration of the LSIG, we select the index corresponding to the largest element on the diagonal of $\left(P^\top_{\overline{A}} \Sigma_\epsilon^{-1}P_{\overline{A}} \right) \Sigma_{Y_{\overline{A}}|Y_{A}}$. That is we solve
\begin{align*}
    i^* = \argmax_i \diag \left(\left(P^\top_{\overline{A}} \Sigma_\epsilon^{-1}P_{\overline{A}} \right) \Sigma_{Y_{\overline{A}}|Y_{A}}, i\right)
\end{align*}
in each iteration. 

For this particular example, the prior covariance matrix $\Sigma_X \in \mathbb R^{d\times d}$ and the noise covariance matrix $\Sigma_\epsilon \in \mathbb R^{n\times n}$ are both generated using the exponential kernel $a\exp(-\vert z_i-z_j \vert/l)$.
In particular, for $\Sigma_{X}$, we set 
\begin{align*}
    z_i &= \frac{i-1}{d-1}\\
    l &= 1/d\\
    a &= 1.
\end{align*}
For $\Sigma_{\epsilon}$, we set 
\begin{align*}
    z_i &= \frac{i-1}{n-1}\\
    l &= 1/n\\
    a &= 0.01.
\end{align*}

\begin{figure}
\centering
\includegraphics[width=0.5\textwidth]{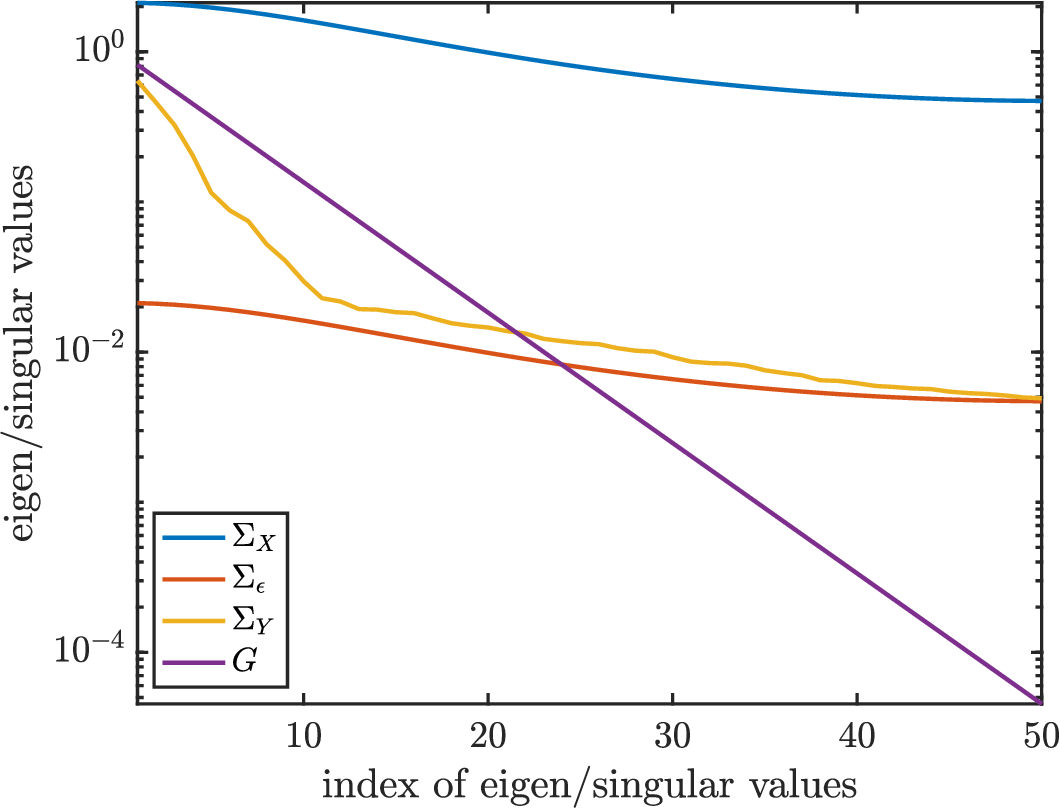}
\caption{The linear Gaussian problem: The spectrum of $G$, $\Sigma_X$, $\Sigma_\epsilon$ and $\Sigma_Y$. }
\label{fig:LG_spectrum}
\end{figure}

\begin{figure}
\centering
\includegraphics[width=0.5\textwidth]{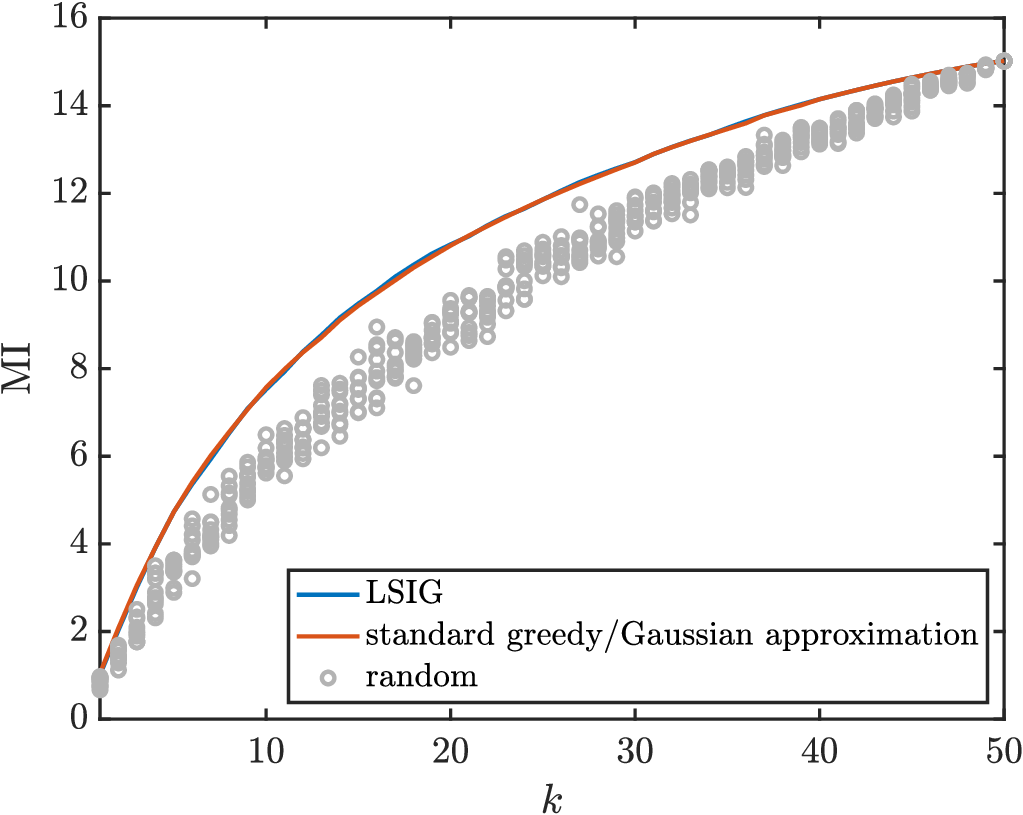}
\caption{The linear Gaussian problem: MI for designs of increasing size, obtained using LSIG, NMC-greedy, and random selection. MI is computed using the closed-form expression in all cases.}
\label{fig:LG_performance}
\end{figure}

\subsection{Epidemic transmission modeling: detailed derivation of $\nabla_{y} \pi_{Y|X}(y|x;t)$ and $\nabla_{y} \log\pi_{Y|X}(y|x;t)$}\label{appendix:epidemic}
We have
\begin{align*}
    \pi_{Y_1, \ldots, Y_n|X}(y_1, \ldots, y_n|x;t_1, \ldots, t_n) = \prod_{i = 1}^n\pi_{Y_i|X}(y_i|x;t_i), 
\end{align*}
then
\begin{align*}
\nabla_{y_i} \pi_{Y_1, \ldots, Y_n|X}(y_1, \ldots, y_n|x;t_1, \ldots, t_n) = \nabla_{y_i} \pi_{Y_i|X}(y_i|x;t_i) \prod_{j \neq i}^n\pi_{Y_j|X}(y_j|x;t_j).
\end{align*}
For each $y_i$,
\begin{align*}
    \nabla_{y_i} \pi_{Y_i|X}(y_i|x;t_i) =& -\frac{\Gamma(N+1)p_i^{y_i}(1-p_i)^{N-y_i}}{\Gamma(y_i+1)\Gamma(N-y_i+1)}\left(-\Psi(N-y_i+1)+\Psi(y_i+1) -\log p_i + \log(1-p_i) \right)\\
    =& -\pi_{Y_i|X}(y_i|x;t_i)\left(-\Psi(N-y_i+1)+\Psi(y_i+1) -\log p_i + \log(1-p_i) \right),
\end{align*}
where $p_i = 1-e^{-xt_i}$, $\Gamma(\cdot)$ is the gamma function and $\Psi(\cdot)$ is the digamma function.

On the other hand, 
\begin{align*}
    \nabla_{y} \log\pi_{Y|X}(y|x;t) = \nabla_{y_i} \log \pi_{Y_i|X}(y_i|x;t_i) = \Psi(N-y_i+1)-\Psi(y_i+1) +\log p_i - \log(1-p_i).
\end{align*}

\subsection{Spatial Poisson model: detailed derivation of $\nabla_{y} \pi_{Y|X}(y|x;b)$ and $\nabla_{y} \log\pi_{Y|X}(y|x;b)$}\label{appendix:poisson}
As in~\ref{appendix:epidemic}, we have
$\nabla_{y_i} \pi_{Y|X}(y|x;b) = \nabla_{y_i} \pi_{Y_i|X_i}(y_i|x_i;b_i) \prod_{j \neq i}^n\pi_{Y_j|X_j}(y_j|x_j;b_j)$. Then since $\pi_{Y_i|X_i}(y_i|x_i;b_i) = \frac{(b_i x_i)^{y_i}}{y_i!}e^{-b_i x_i}$, we have
\begin{align*}
    \nabla_{y_i} \pi_{Y_i|X_i}(y_i|x_i;b_i) = \frac{e^{-b_i x_i}(b_i x_i)^{y_i} \left(\log(b_ix_i)-\Psi(y_i+1) \right)}{y_i!}
\end{align*}
for all $i$, and $\Psi(\cdot)$ is the digamma function. Furthermore, 
\begin{align*}
    \nabla_{y} \log\pi_{Y|X}(y|x;b) = \nabla_{y_i} \log \pi_{Y_i|X_i}(y_i|x_i;b_i) = \log(b_ix_i) -\Psi(y_i+1).
\end{align*}

\begin{figure}
\centering
\includegraphics[width=0.5\textwidth]{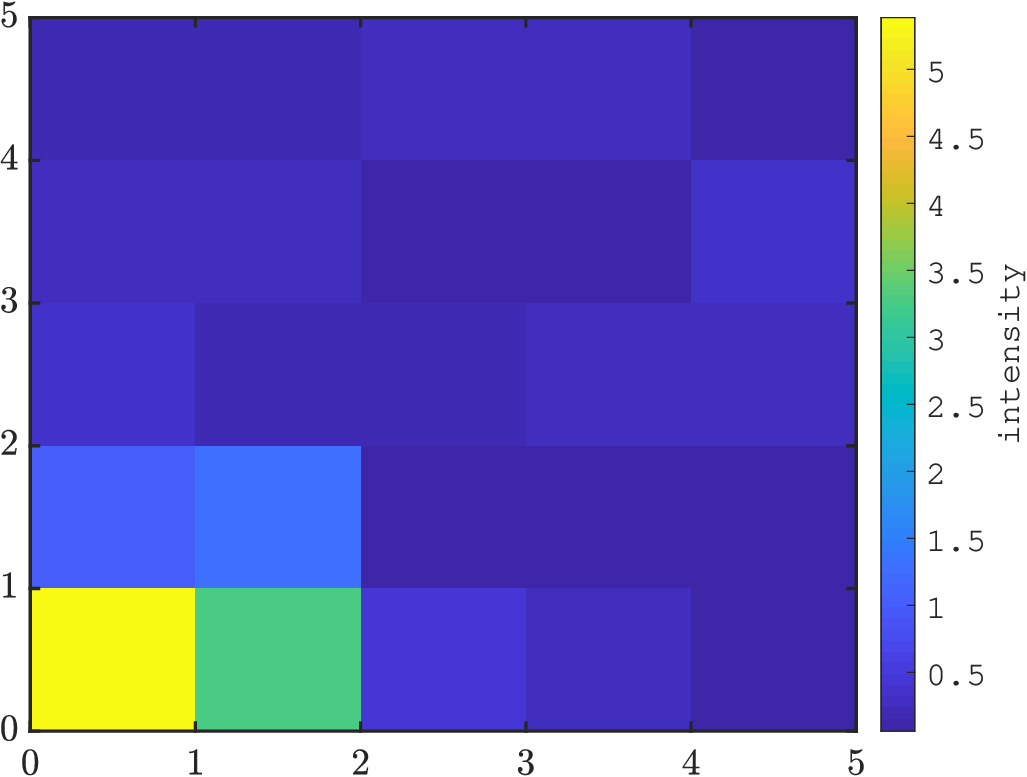}
\caption{The spatial Poisson process: an instance of the underlying intensity field}
\label{fig:poisson_intensity}
\end{figure}

\begin{figure}[h]
\centering
\includegraphics[width=0.5\textwidth]{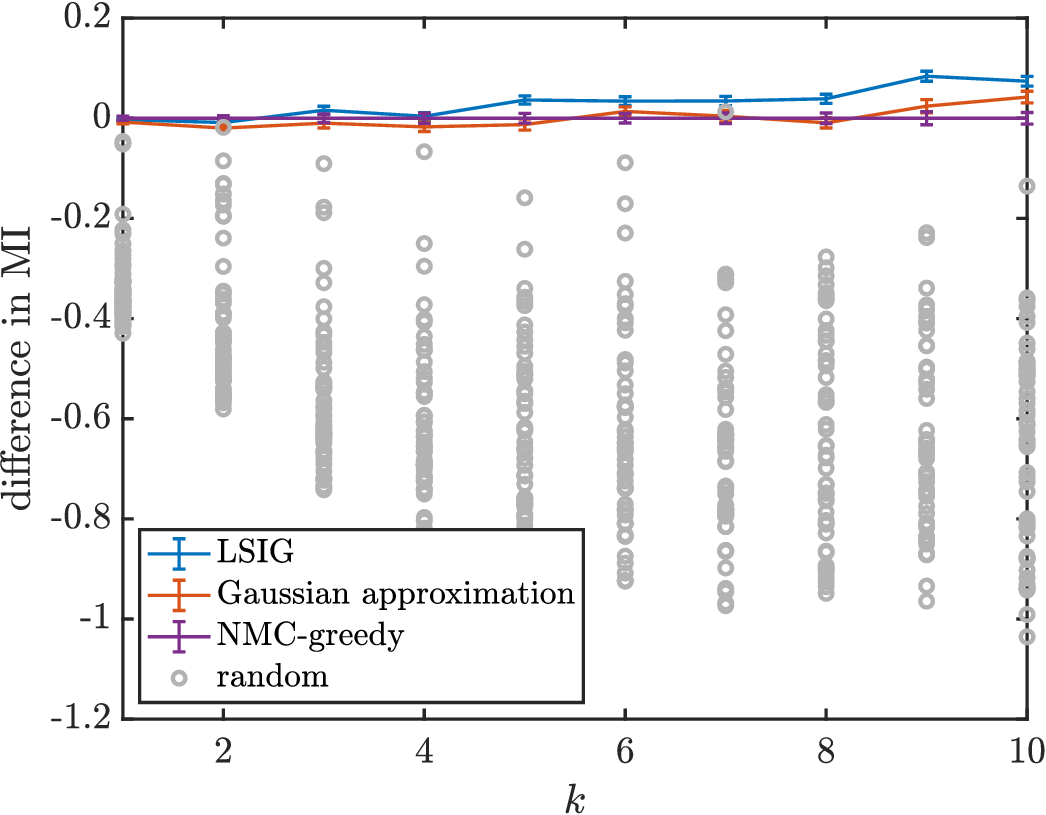}
\caption{Spatial Poisson process: differences in MI for designs obtained with \{LSIG, Gaussian approximation, random selection\} and with NMC-greedy (i.e., results obtained using NMC-greedy set the zero value at each $k$). MI values for the vertical axis are computed using NMC with $10000$ inner samples and $1000$ outer samples. Error bars show one standard error, computed over $50$ trials.}
\label{fig:poisson_MI_diff_with_random}
\end{figure}

\end{document}